%% file: main.tex
\documentclass[runningheads]{llncs}

\usepackage{eccv}

\usepackage{eccvabbrv}

\usepackage{graphicx}
\usepackage{booktabs}

\usepackage[accsupp]{axessibility}  

\usepackage{hyperref}

\usepackage{orcidlink}

\usepackage{array}      
\usepackage{booktabs}   
\usepackage{multirow}   
\usepackage{rotating}   
\usepackage{graphicx}   
\usepackage{boldline}
\usepackage{comment}
\usepackage{makecell}   
\usepackage{xcolor}     
\usepackage{amsmath}    
\usepackage{caption}    
\usepackage{wrapfig}
\usepackage{float}
\usepackage{adjustbox}

\begin{document}

\title{CL-Anomaly: Layer-Adaptive Mixture-of-Experts with Multimodal Large Language Model for Continual Learning in Anomaly Detection} 

\titlerunning{CL-Anomaly: Layer-Adaptive MoE with MLLM for CL in AD}

\author{
Wen Dong\inst{1} \and
Zhao Wang\inst{2}\textsuperscript{\textdagger} \and
Shuangqing Zhang\inst{1}\and
Kai Sun\inst{2} \and
Ben Li\inst{2} \and  \\[4pt] 
Guo-Sen Xie\inst{3} \and
Caifeng Shan\inst{1} \and
Fang Zhao\inst{1}\textsuperscript{\textdagger}
}

\makeatletter
\def\daggerfootnote{\gdef\@thefnmark{\textdagger}\@footnotetext}
\makeatother

\authorrunning{W.~Dong et al.}

\institute{
Nanjing University, Nanjing, China\\
 \and
China Mobile Zijin Innovation Institute, Nanjing, China\\
\and
Nanjing University of Science and Technology, Nanjing, China \\
\email{\{wdong, shuangqing\}@smail.nju.edu.cn} \\
\email{\{wangzh8, sunk, liben\}@js.chinamobile.com} \\ 
\email{\{gsxiehm, caifeng.shan, zhaofang0627\}@gmail.com}
}

\maketitle
\input{sec/0_abstract}    
\daggerfootnote{Corresponding authors.}

\input{sec/1_introduction}
\input{sec/2_related}
\input{sec/3_preliminary}
\input{sec/4_method}
\input{sec/5_experiments}

\section{Conclusion}
In this paper, we propose the CL-Anomaly framework, which first introduces continual learning with MLLM into anomaly detection task. We design a dual-path architecture combining parameter decoupling and shared experts to balance task-specific knowledge retention and cross-task knowledge transfer. The parameter spaces of the private expert PrivLoRA are isolated, with only the task-specific low-rank matrices updated. Shared experts enable multiplexing of multi-task knowledge through sparse gating within a unified representation space. We further observed that experts at different layers show distinct task preferences, motivating our Layer-Adaptive Knowledge Transfer strategy, which automatically identifies key experts at each layer with zero additional cost and enhances their contribution through dynamic momentum merging. This approach preserves critical information and promotes knowledge transfer, effectively mitigating catastrophic forgetting. Extensive experiments demonstrate that CL-Anomaly consistently outperforms SOTA methods across class-incremental, cross-domain, and cross-modal continual learning scenarios in anomaly detection tasks.
\input{sec/7_acknowledgements}

\bibliographystyle{splncs04}

\bibliography{main}
\input{Supplementary_Material/appendix}

\end{document}

%% file: sec/0_abstract.tex
\begin{abstract}
Multimodal Large Language Models (MLLMs) excel in diverse vision tasks, but full-parameter retraining is computationally expensive as real-world knowledge evolves. Existing continual learning methods often suffer from semantic entanglement in parameter spaces across tasks, impeding the continuous deployment of models. This challenge is especially pronounced in Anomaly Detection (AD), which exhibits triple heterogeneity across modalities, domains, and defect scale variability, significantly complicating multi-task knowledge transfer. In this paper, we propose CL-Anomaly, a parameter-efficient fine-tuning framework based on an isolation-sharing collaboration to enable continual learning for anomaly detection with MLLMs. We introduce the task-private expert PrivLoRA, which physically isolates task-specific subspaces in the parameter space to prevent semantic entanglement of anomaly knowledge in diverse scenarios. The Layer-Adaptive Shared Experts maintain cross-task representations within a unified feature space, enabling knowledge sharing between previous and new tasks. Furthermore, we propose a Layer-Adaptive Knowledge Transfer strategy that automatically selects and dynamically updates the layer-wise key shared experts of each task via a momentum-based mechanism, promoting effective knowledge transfer across related anomaly detection tasks. Extensive experiments across three continual learning scenarios for anomaly detection, including class-incremental, cross-domain, and cross-modal, demonstrate that CL-Anomaly outperforms state-of-the-art methods. Code is available at \url{https://github.com/WenDongyp/CL-Anomaly}.
  \keywords{Anomaly Detection \and Continual Learning \and Multimodal Large Language Model}
\end{abstract}

%% file: sec/1_introduction.tex
\section{Introduction}
Anomaly detection (AD) aims to identify samples that deviate from normal data distribution and has been widely applied in areas such as industrial quality control\cite{roth2022towards,cao2023anomaly,tien2023revisiting,he2024mambaad} and medical imaging diagnosis\cite{wei2018anomaly,zhang2024mediclip,huang2024adapting,zhao2021anomaly}. Conventional methods typically provide only anomaly scores, offer no user interaction, and are often restricted to a single modality (e.g., 2D or 3D), as well as specific task domains. In contrast, multimodal large language models (MLLMs)\cite{li2024llava,achiam2023gpt,chen2024internvl,bai2025qwen2,fu2025vita} inherently support cross-modal semantic understanding and human–computer interaction, allowing instruction prompts to guide the model in generating desired responses while handling inputs across multiple modalities and domains. Consequently, leveraging MLLMs for anomaly detection represents a promising direction. However, real-world anomaly detection is complex and evolving, making per-pipeline model training costly and impractical. Moreover, historical data privacy, commonly encountered in practical AD systems, restricts the application of many existing anomaly detection paradigms. Equipping MLLMs with continual learning capabilities enables incremental adaptation to new categories and domains, offering an effective solution. Hence, advancing continual learning for anomaly detection with MLLMs is of critical significance.

\begin{figure*}[t]
\centering
  \includegraphics[width=\linewidth]
  {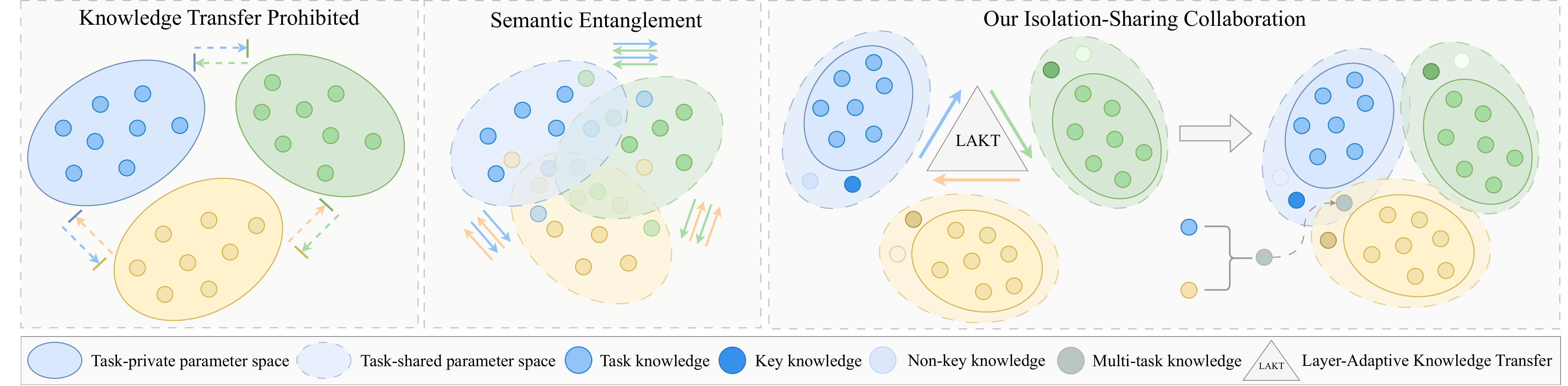}
  \caption{(Left) Parameter isolation methods prohibit cross-task knowledge transfer, preventing knowledge complementarity. (Middle) Parameter sharing methods often lead to semantic entanglement, where parameter spaces of multiple tasks overlap and conflict. (Right) Our isolation–sharing collaboration framework maintains task-specific knowledge within isolated parameter spaces while enabling cross-task knowledge transfer through shared spaces. Furthermore, we introduce a layer-adaptive knowledge transfer strategy that preserves layer-wise key knowledge during parameter updates and facilitates the formation of multi-task knowledge, while non-key knowledge gradually fades throughout continual learning.}
  \label{fig:motivation}
\end{figure*}

\begin{figure}[ht]
    \centering
    \begin{subfigure}[b]{0.32\linewidth}
        \centering
        \includegraphics[width=\linewidth]{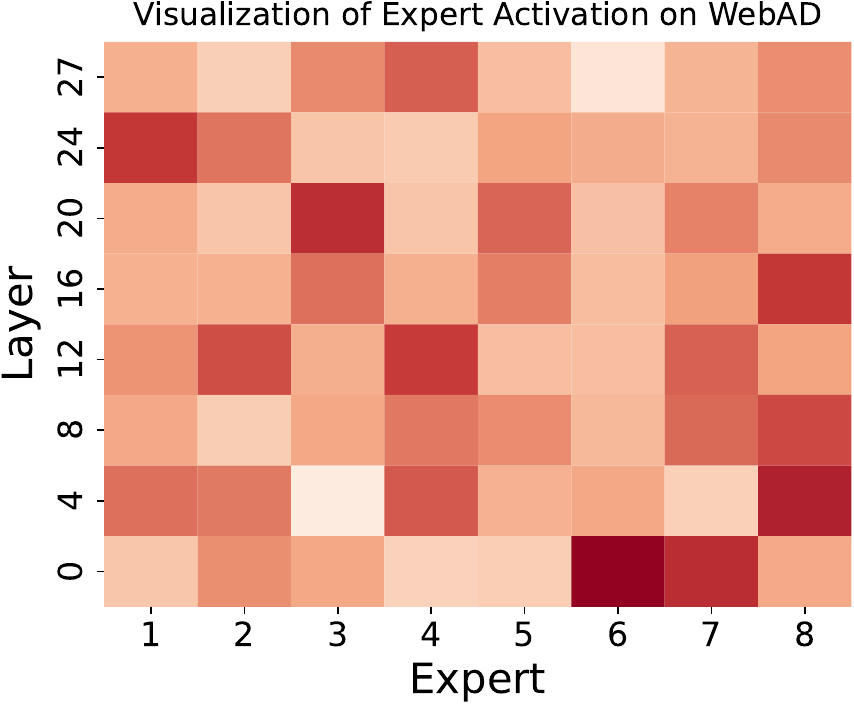}
        \caption{LLava-OV-7B}
    \end{subfigure}
    \hfill
    \begin{subfigure}[b]{0.32\linewidth}
        \centering
        \includegraphics[width=\linewidth]{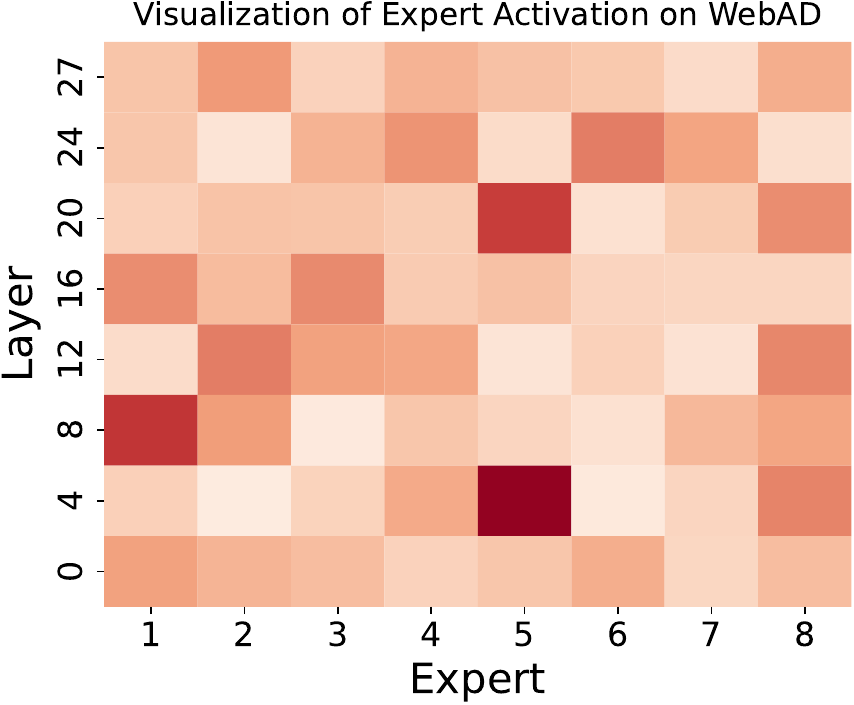}
        \caption{Qwen3-VL-2B}
    \end{subfigure}
    \hfill
    \begin{subfigure}[b]{0.32\linewidth}
        \centering
        \includegraphics[width=\linewidth]{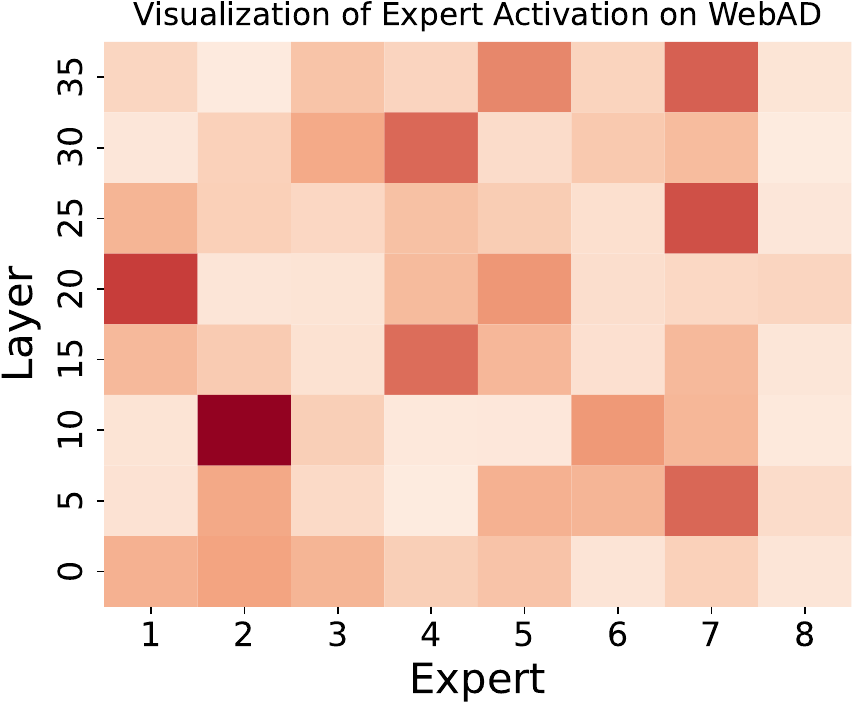}
        \caption{Qwen3-VL-8B}
    \end{subfigure}

    \begin{subfigure}[b]{0.32\linewidth}
        \centering
        \includegraphics[width=\linewidth]{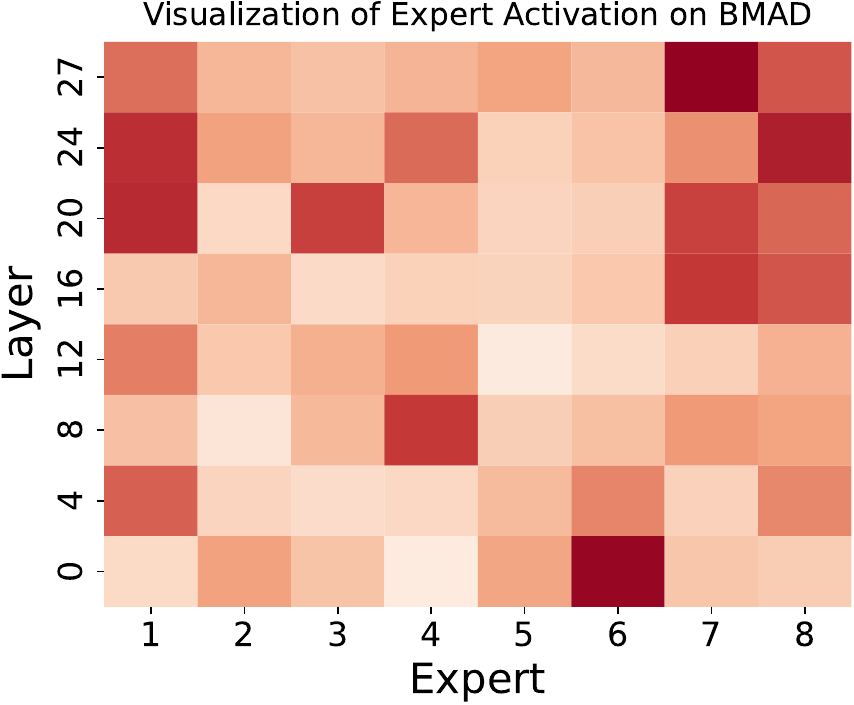}
        \caption{LLava-OV-7B}
    \end{subfigure}
    \hfill
    \begin{subfigure}[b]{0.32\linewidth}
        \centering
        \includegraphics[width=\linewidth]{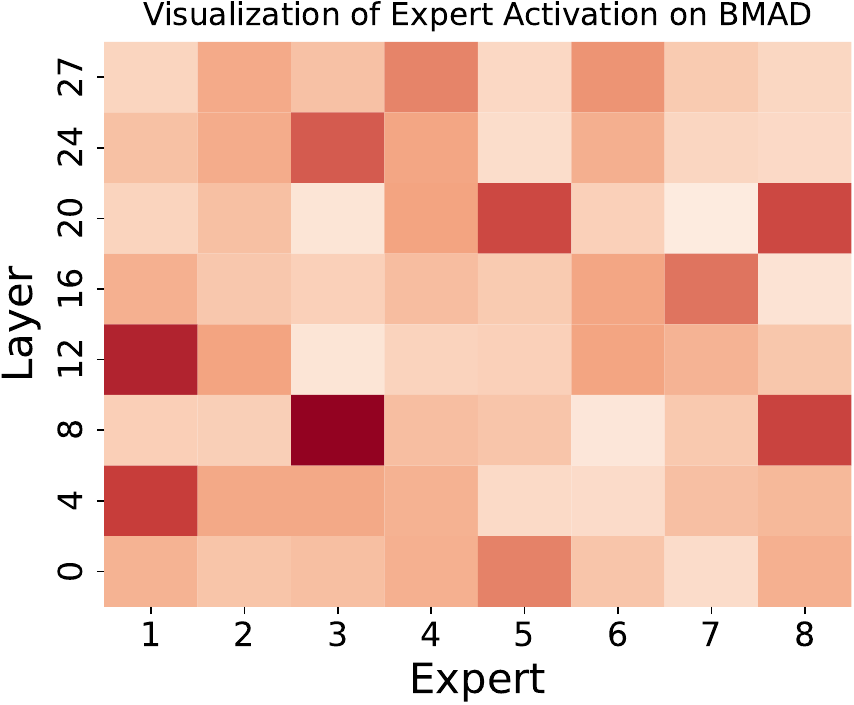}
        \caption{Qwen3-VL-2B}
    \end{subfigure}
    \hfill
    \begin{subfigure}[b]{0.32\linewidth}
        \centering
        \includegraphics[width=\linewidth]{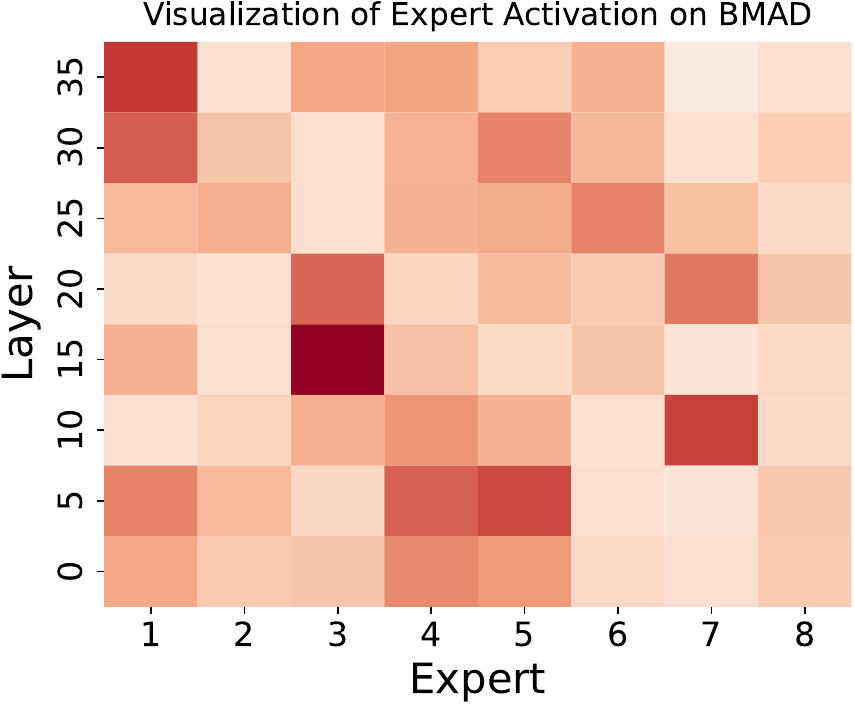}
        \caption{Qwen3-VL-8B}
    \end{subfigure}
    \caption{Heatmaps of expert activation across different layers in the LoRA-based MoE architecture on WebAD and BMAD datasets. Layer-wise experts exhibit varying activation preferences across tasks, which is widely observed across MLLMs of diverse scales and architectures, highlighting the importance of differentiating layer-wise experts based on their task preferences.}
    \label{fig:activity}
\end{figure}

Parameter-efficient fine-tuning\cite{houlsby2019parameter,li2021prefix,lester2021power,hu2022lora, zhao2025mllm} is the most prevalent continual learning paradigm for MLLMs, introducing few trainable parameters for rapid task adaptation while keeping pretrained weights frozen. Nevertheless, it inevitably faces catastrophic forgetting, the classical challenge in continual learning, which is typically caused by semantic entanglement across tasks or the absence of knowledge transfer mechanisms. Recently, many efforts have been made to mitigate catastrophic forgetting. MoELoRA\cite{chen2024coin}, SMoLoRA\cite{wang2025smolora}, and CL-MoE\cite{huai2025cl} update parameters within a unified representation space, enabling positive knowledge transfer across similar tasks. O-LoRA\cite{wang2023orthogonal}, MoA\cite{feng2024mixture}, and HiDe-LLaVA\cite{guo2025hide} isolate parameter spaces for individual tasks, effectively preventing semantic entanglement. However, relying solely on either approach entails inherent limitations: shared parameters risk conflict when tasks span diverse domains, while isolated parameters impede knowledge transfer across tasks, as illustrated in \cref{fig:motivation} (Left, Middle).

In this paper, we propose CL-Anomaly to bridge the research gap in continual learning for anomaly detection with MLLMs. As shown in \cref{fig:motivation} (Right), CL-Anomaly incorporates task-private and task-shared paths, with Layer-Adaptive Knowledge Transfer executed on the shared parameter space. In the task-private path, each task employs a dedicated LoRA expert, termed PrivLoRA, which updates only within its own task. Fully isolating the parameter spaces of PrivLoRAs inherently mitigates semantic interference in anomaly knowledge across modalities and domains. The task-shared path enables cross-task knowledge transfer by jointly optimizing multiple LoRAs in a unified parameter space. Together, the two paths effectively mitigate catastrophic forgetting. Moreover, we further observe that in the LoRA-based MoE architecture, the activity of individual LoRA experts $e$ at each layer $l$ varies across different tasks $t$, as illustrated in \cref{fig:activity}. Based on this observation, we propose Layer-Adaptive Knowledge Transfer, a simple yet efficient strategy to preserve the knowledge of layer-wise key shared experts. During training, it automatically identifies highly active experts for the current task at zero extra cost and prioritizes their information when updating shared parameters via dynamic momentum merging, thereby preserving core anomaly knowledge while facilitating knowledge integration across related anomaly detection tasks.
In summary, our contributions are as follows:

\begin{itemize}
    \item We introduce CL-Anomaly, which adopts a \textit{private-shared} dual-path fine-tuning framework to prevent semantic entanglement while enabling knowledge transfer. To our knowledge, we are the first to bring the continual learning paradigm of MLLMs into anomaly detection.

    \item We found significant differences in task preferences among experts at different layers and propose a Layer-Adaptive Knowledge Transfer strategy to adaptively retain layer-wise key expert knowledge while facilitating cross-task knowledge transfer.

    \item Extensive experiments on three continual learning scenarios for anomaly detection, including class-incremental, cross-domain, and cross-modal, show that CL-Anomaly effectively mitigates catastrophic forgetting and outperforms current state-of-the-art methods.
    
\end{itemize}

%% file: sec/2_related.tex
\section{Related Work}
\noindent\textbf{Anomaly Detection.}
Early anomaly detection methods relied mainly on reconstruction error\cite{liu2023anomaly,lin2023latent,givnan2022anomaly} or feature matching\cite{roth2022towards,wang2021student,liu2023simplenet} to detect anomalies, following a \textit{one-class-one-model} training paradigm, which is highly impractical in real-world scenarios. In recent years, researchers have focused on few-shot\cite{li2024promptad,jeong2023winclip,tao2025kernel} and zero-shot\cite{zhou2023anomalyclip,cao2024adaclip,dong2025region} learning paradigms, leveraging visual-language models with strong generalization capabilities (e.g., CLIP) to calculate anomaly scores by comparing visual and textual feature representations, but their detection capability is often constrained by limited exposure to the target dataset. Real-world scenarios, including the continual emergence of new product categories in industry and cross-domain transfer from industrial inspection to medical imaging, exemplify the significance of continual anomaly detection, yet few studies have investigated it. Accordingly, we propose the CL-Anomaly framework, enabling efficient continual learning for anomaly detection across both class and domain incremental scenarios.

\noindent\textbf{Multimodal Large Language Models.}
With the rapid development of large language models (LLMs)\cite{bai2023qwen,touvron2023llama}, growing attention has been devoted to equipping them with multimodal learning capabilities, thereby extending their applications to complex scenarios. Some studies integrate image encoders with modality connectors into LLMs to enable visual understanding, gradually expanding to additional modalities such as video and audio. These multimodal large language models (MLLMs) demonstrate outstanding capabilities in instruction following\cite{liu2023visual,gao2023llama}, knowledge reasoning\cite{chen2024lion,zheng2024thinking,gao2025interleaved} and various vision downstream tasks\cite{lin2024video,wu2024visionllm,wang2026temporal}. However, research on MLLMs for anomaly detection remains scarce. This paper aims to leverage MLLMs for continual learning in anomaly detection, providing a novel insight and method for multimodal anomaly detection.

\noindent\textbf{Continual Learning.}
In real-world, the continuous evolution of product categories and task types is widespread. Accordingly, MLLMs are expected to dynamically acquire new knowledge while robustly retaining previously learned information. Recent continual learning methods can be mainly categorized into replay-based, regularization-based, and parameter-efficient fine-tuning methods. Replay-based methods\cite{zhu2021prototype,rolnick2019experience,lopez2017gradient}, preserve a shared pre-trained representation space by storing and replaying knowledge from previous tasks. However, as the accumulated knowledge grows, these methods incur substantial storage overhead and are constrained in anomaly detection scenarios involving data privacy.
Regularization-based methods\cite{kirkpatrick2017overcoming,dhar2019learning,li2017learning} preserve the stability of previous task parameters during incremental learning through explicit constraints on critical model parameters or the loss function. Parameter-efficient fine-tuning methods achieve rapid adaptation to new tasks by updating a limited number of additional parameters while keeping the pre-trained backbone frozen to preserve the model’s original knowledge. MoELoRA\cite{chen2024coin} learns diverse knowledge representations through the collaboration of multiple LoRA experts. CL-MoE\cite{huai2025cl} employs a dynamic expert update strategy, achieving a trade-off between plasticity and stability. But the absence of parameter decoupling among experts renders CL-MoE susceptible to semantic entanglement. Our method combines parameter-decoupled private experts with shared experts, maintaining task-specific knowledge while promoting cross-task knowledge transfer. HiDe-LLaVA\cite{guo2025hide} introduces task-specific expansion and task-general fusion but treats all non-top-layer LoRA experts uniformly, neglecting the layer-wise differences in expert task sensitivity. Our method explicitly accounts for layer-wise expert task preferences, dynamically identifying layer-wise key experts salient to the current task and  effectively preserving their knowledge during transfer.

%% file: sec/3_preliminary.tex
\section{Preliminary}
\textbf{Continual Learning Setup.}
Our work focuses on continual learning for anomaly detection with MLLM, aiming to sequentially learn tasks while mitigating catastrophic forgetting. Given $T$ tasks, continual instruction tuning sequentially accesses the data of each task
$\mathcal{D}_t = \{V_i^t, Q_i^t, A_i^t\}_{i=1}^{N^t},~t \in \{1,2,\dots,T\},$ where $N^t$ denotes the data size of task $t$. The MLLM is tasked with learning the current task $t$ while preserving the knowledge from previous tasks $t-1$.

\noindent\textbf{Low-Rank Adaptation.}
Low-Rank Adaptation (LoRA) is a parameter-efficient fine-tuning method. By freezing the pretrained weights $W_0 \in \mathbb{R}^{d \times k}$ and introducing a trainable low-rank update $\Delta W = BA$, where $B \in \mathbb{R}^{d \times r}$, $A \in \mathbb{R}^{r \times k}$, and the rank $r \ll \min(d, k)$, it adapts the model to new tasks while significantly reducing the number of updated parameters. It is typically injected into the linear projection layers of the self-attention and feed-forward networks. The forward pass with LoRA can be expressed as: 
\begin{equation}
f(x) = W_{0}x+\Delta W x =W_{0} x + \frac{\alpha}{r}\,B\,(A x)
\end{equation}
where $\alpha$ denotes the scaling factor controlling the magnitude of the low-rank update.

%% file: sec/4_method.tex
\begin{figure*}[t]
\centering

  \includegraphics[width=\linewidth ]
  {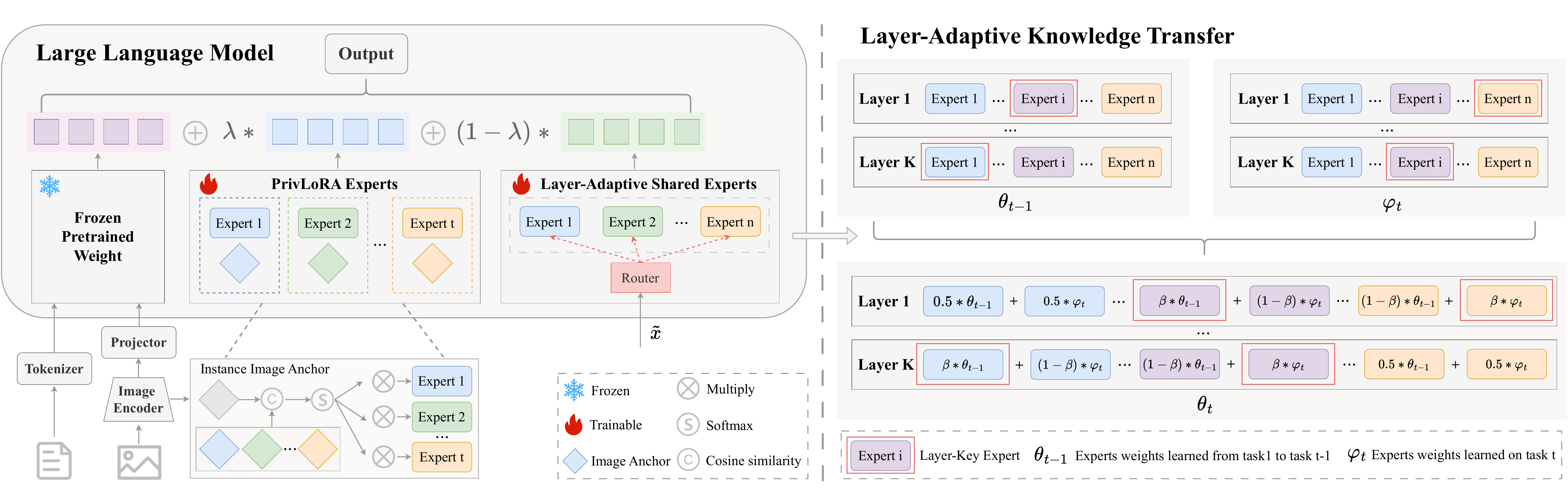}
  \caption{Overview of our method. (Left) CL-Anomaly consists of parallel task-private and task-shared paths. PrivLoRA is updated only within its task, extracting visual anchors with a multimodal encoder and assigning expert weights by similarity to task anchors. The task-shared path, termed Layer-Adaptive Shared Experts, dynamically selects experts with a gating function and updates parameters using a layer-adaptive strategy. (Right) Layer-Adaptive Knowledge Transfer: identifies key layer-wise experts from gating probability vectors at zero extra cost and prioritizes key expert representations during dynamic momentum merging.}
  \label{fig:framework}
\end{figure*}

\section{Method}

\subsection{Framework Overview}
In this work, we propose CL-Anomaly, a dual-path mixture-of-experts framework for continual learning in anomaly detection. The overall architecture is illustrated in \cref{fig:framework}. CL-Anomaly introduces parallel task-private and task-shared expert paths, enabling simultaneous semantic decoupling and knowledge transfer. During training, only current PrivLoRA and task-shared experts are updated, while other private experts remain frozen. At the inference stage, the model combines adapted increments from both expert paths with frozen weights output:
\begin{equation}
h = W_{0}x+\lambda f_{p}(x)+(1-\lambda)f_{s}(x)
\end{equation}
where $\lambda$ denotes the weighting coefficient, fixed at $0.5$. $f_{p}(\cdot)$ and $f_{s}(\cdot)$ denote the outputs of the private and shared paths, with further discussion in Sections~\ref{sec:priv} and~\ref{sec:shared}.

To further mitigate catastrophic forgetting, we propose a layer-adaptive knowledge transfer strategy that identifies key experts layer by layer and dynamically preserves their knowledge during parameter updates, enabling efficient layer-wise task-critical knowledge transfer.

\subsection{PrivLoRA Experts}
\label{sec:priv}
Conventional LoRA-based fine-tuning approaches are inherently susceptible to semantic entanglement in multi-task continual learning. To overcome this limitation, inspired by HiDe-LLaVA\cite{guo2025hide}, we further introduce task-private PrivLoRA experts. Each task is assigned a dedicated PrivLoRA expert, receiving gradient updates exclusively during the training of the corresponding task. PrivLoRA experts maintain parameter spaces fully isolated from other tasks, preventing cross-task interference in the shared representation space and eliminating semantic entanglement at its source. We use the visual encoder of LLaVA to extract the visual anchor $anchor_t$ for task $t$ during training:
\begin{equation}
\label{eq:1}
anchor_{t}=\frac{1}{N_{t}}\sum_{i=1}^{N_{t}}(\frac{1}{M}\sum_{m=1}^{M} [f_v(x_i^t)]_m)
\end{equation}
where $x^t_i$ denotes the input image instance $i$ of task $t$, $f_v$ is the visual encoder of LLaVA, and $f_v(x) \in \mathbb{R}^{B \times M \times C}$ denotes the patch features of input $x$, with $B$ the batch size, $M$ the number of patch tokens, and $C$ the feature dimension.

The PrivLoRA expert appropriate for the input image $x_{I}$ is selected based on the cosine similarity between its instance visual anchor $anchor_{I}$ and the stored task anchors. Specifically, we quantify the correlation between the input instance visual anchor $anchor_{I}$ and the stored anchor of task $t$ as $\mu_{t}$:
\begin{equation}
\label{eq:2}
\mu_{t}= \frac{\exp\bigl(\mathrm{Cos}({anchor}_I,\, {anchor}_{t})\bigr)}{\sum_{j=1}^{T}\exp\bigl(\mathrm{Cos}({anchor}_I,\, {anchor}_{j})\bigr)}, \quad t=1,\dots,T
\end{equation}
where $\mathrm{Cos}(.,.)$ represents the cosine similarity. The output of the PrivLoRA experts path is given as follows:
\begin{equation}
\label{eq:3}
f_{p}(\tilde{x_{I}})=\sum_{i=1}^{t}\mu_{i}P_{i}(\tilde{x}_{I})
\end{equation}
where $\tilde{x_{I}}$ denotes the intermediate representation of the input instance $x_{I}$ within the large language model block, and $P_{i}(\cdot)$ represents the $i$-th PrivLoRA expert.

\subsection{Layer-Adaptive Shared Experts}
\noindent\textbf{Task-Shared Experts.}
\label{sec:shared}
Although physically isolated task-private experts effectively eliminate semantic entanglement, the complete decoupling of parameters simultaneously hinders knowledge transfer across tasks. Prior studies\cite{lin2022beyond,qiao2024learn,piao2024federated} have demonstrated that knowledge transfer supports the progressive accumulation of information, with new task knowledge complementing and reinforcing the representations of previous tasks, especially when there are intrinsic semantic or distributional correlations, e.g., in class-incremental learning under a shared imaging environment. We thus employ a task-shared expert path to complement PrivLoRA, addressing its constraints in cross-task knowledge transfer. The task-shared path adopts a vanilla mixture-of-experts LoRA architecture, treating each LoRA as an individual expert network. In this architecture, the router mapping the input $x$ to expert assignment weights $\omega$:
\begin{equation}
\label{eq:4}
\omega_{e}(x) = \frac{\exp(g_e \cdot x)}{\sum_{j=1}^{n} \exp(g_j \cdot x)}, \quad e = 1,\dots,n
\end{equation}
where $g_{e}$ denotes the trainable gating parameter of expert $e$ and $n$ is the number of shared experts. The output of the task-shared path is obtained by weighting the experts accordingly:
\begin{equation}
f_{s}(\tilde{x})=\sum_{e=1}^{n}\omega_{e}(\tilde{x})\cdot F_{e}(\tilde{x})
\end{equation}
where $F_{e}(\cdot)$ denotes the shared LoRA expert $e$.

\noindent\textbf{Layer-Adaptive Knowledge Transfer.}
We further observe significant differences in task sensitivity among experts $e$ at each layer $l$ for different tasks $t$ within the LoRA-based MoE architecture, with individual experts showing varying activation in fine-grained tasks and global understanding. Therefore, we introduce a Layer-Adaptive Knowledge Transfer strategy to identify key layer-wise LoRA experts and enhance cross-task knowledge propagation. Specifically, we compute the task relevance score $S$ of shared experts per layer during training by accumulating gating probability vectors. 
\begin{equation}
\label{eq:5}
S_t^{(l,e)} = \frac{1}{|\mathcal{D}_t|} 
\sum_{x \in \mathcal{D}_t} 
\omega_{l,e}(x) \cdot 
\mathbb{I}\!\left[e = \operatorname*{argmax}_{j \in \{1,\dots,n\}} \omega_{l,j}(x)\right]
\end{equation}
where $\omega_{l,e}(x)$ denotes the routing probability of input sample $x$ to expert $e$ at layer $l$. The top-$K$ shared experts with the highest scores $S_t^{(l,e)}$ are designated as key LoRA experts $e_{t,l}^{*}$, remaining consistently active at layer $l$ for task $t$ and demonstrating strong task relevance. This is a simple yet efficient strategy that selects key experts with minimal computational overhead, without requiring additional forward or backward propagation. Let $E_{t}^{*}=\{ \,e_{t,l}^* \mid l = 1,2,\dots,L \,\}$ denote the set of layer-wise key experts for task $t$, where $L$ is the number of layers in the LLM. We partition all shared experts into three mutually exclusive subsets: 
$E_t^{only} = \{ e \in E_t^* \mid e \notin E_{t-1}^* \}$, the experts key only to task $t$; 
$E_{t-1}^{only} = \{ e \in E_{t-1}^* \mid e \notin E_t^* \}$, the experts key only to task $t-1$; 
and 
the remaining experts. The transfer coefficient of shared expert $e$ at layer $l$ is defined as:
\begin{equation}
\beta_{l,e} =
\begin{cases}
\beta, & \text{if }e \in E_t^{only} \\
1-\beta, & \text{if }e \in E_{t-1}^{only} \\
0.5, & \text{otherwise }
\end{cases}
\end{equation}
Let $\theta_{t-1}$ denote the shared expert parameters learned by the model from tasks $1$ to $t-1$, and let $\varphi_{t}$ represent the parameters initialized with $\theta_{t-1}$ and fine-tuned on task $t$. We perform layer-adaptive momentum merging of $\theta_{t-1}$ and $\varphi_{t}$ to enable efficient knowledge transfer across tasks, thereby maximizing the update of key knowledge for the new task while retaining knowledge from previous tasks. The process can be expressed as:
\begin{equation}
\theta_{t}^{(l,e)} = (1-\beta_{l,e})\cdot\theta_{t-1}^{(l,e)}+\beta_{l,e}\cdot \varphi^{(l,e)}_{t}
\end{equation}
where $\theta_{t}$ denotes the updated shared expert parameters for task $t$. The layer-adaptive knowledge transfer strategy retains key expert knowledge across all layers for each task, facilitating knowledge transfer while effectively mitigating catastrophic forgetting.

%% file: sec/5_experiments.tex
\section{Experiments}

\subsection{Experimental Setup}
\noindent\textbf{Datasets.}
We conduct experiments on Anomaly-Instruct-125K\cite{xu2025towards}, the largest anomaly-detection-related visual instruction question-answering dataset to date.
Spanning 2D to 3D data and covering industrial and medical domains, Anomaly-Instruct-125K is composed of classical datasets including MVTec-AD\cite{bergmann2019mvtec}, MVTec-3D\cite{bergmann2021mvtec}, Real3D-AD\cite{liu2023real3d}, Anomaly-ShapeNet\cite{li2024towards}, and BMAD\cite{bao2024bmad}, along with WebAD\cite{xu2025towards}, an in-the-wild multi-class image dataset. We create three subsets from WebAD based on object categories, each containing 35 classes, to evaluate model performance in class-incremental learning scenarios. BMAD, a medical-domain dataset, is used to assess model performance under cross-domain continual learning. Additionally, 3D datasets are employed to evaluate model capability in continual learning across modalities. See Appendix ~\ref{apd:dataset} for more dataset details.

\noindent\textbf{Evaluation Metrics.}
Following prior continual learning works\cite{ge2025dynamic,chen2024coin,huai2025cl,guo2025hide}, we evaluate continual learning performance using $Avg$, $Last$, and $BWT$ metrics based on per-task accuracy across stages. Given the inherent class imbalance in anomaly detection datasets, we further report $Accuracy$, $Precision$, $Recall$, and $F1\mathchar`\-score$ to ensure a fair and comprehensive evaluation of continual anomaly detection performance. Detailed definitions and metric selection criteria are provided in Appendix~\ref{apd:c1}.

\noindent\textbf{Baselines.}
We compare CL-Anomaly with existing continual learning baseline methods. LoRA\cite{hu2022lora} is a standard parameter-efficient fine-tuning approach that is widely adopted for MLLM adaptation. MoELoRA\cite{chen2024coin} leverages a mixture of LoRA experts to learn diverse knowledge representations. CL-MoE\cite{huai2025cl} employs dynamic updates of a mixture of experts to balance expert workload. HiDe-LLaVA\cite{guo2025hide} combines non-top-layer experts while preserving the independence of top-layer experts. We also report performance under zero-shot and multi-task fine-tuning settings, where all tasks are learned simultaneously in the multi-task setup without  task-ordering constraints.
The detailed of each method can be found in Appendix~\ref{apd:c2}.

\noindent\textbf{Implementation details.}
We use LLaVA-OV-7B\cite{li2024llava} as the base MLLM, keeping all its parameters frozen, and embed LoRA experts into the linear layers of the language model following the standard LoRA fine-tuning protocol employed by LLaVA. We set the number of shared experts $n$ to $4$, with $k = 1$ key expert per layer, the LoRA rank $r = 36$, the scaling factor $\alpha = 72$, and the momentum merging weight $\beta = 0.75$. All tasks are trained for 1 epoch using the AdamW\cite{loshchilov2017decoupled} optimizer, with a learning rate of $1\times10^{-5}$ and a warm-up ratio of $0.03$. For fair comparison, all baseline methods use the same base model and fine-tuning strategy with carefully tuned hyperparameters. All experiments are conducted on $8$ RTX A6000 GPUs. Further details are provided in Appendix~\ref{apd:c3}.

\begin{table*}[ht]
    \centering
    \caption{Comparison of our method with various approaches on continual learning for anomaly detection. The best results are highlighted in \textbf{bold}, and the second best results are \underline{underlined}.}
    \resizebox{0.95\linewidth}{!}{
    \begin{tabular}
        {clccccccccc}
        \toprule
           & {\hspace{1em}} Method & WebAD-1 & WebAD-2 & WebAD-3 & Mvtec-AD & BMAD & Mvtec-3D & Real3D-AD & Anomaly-Shapenet & \textit{Average} \\
  
        \midrule

            & {\hspace{1em}}Zero-shot & 86.17 & 86.44 & 84.54 & 92.05 & 61.53 & 69.06 & 55.58 & 56.62 & 73.99 \\   
            & {\hspace{1em}}Multi-task & 93.41 & 92.47 & 93.48 & 97.73 & 92.59 & 83.92 & 77.69 & 75.71 & 88.38 \\  
            \midrule
            \multirow{5}{*}{\rotatebox{90}{\textbf{Avg}}} 
            &{\hspace{1em}}LoRA & 88.59 & 87.63 & 86.13 & 94.14 & 73.97 & \underline{80.87} & \underline{72.91} & \textbf{76.36} & \underline{82.58} \\
            & {\hspace{1em}}MoELoRA  & 89.79 & 88.85 & 88.22 & 94.94 & 74.59 & 77.47 & 62.65 & 71.82& 81.04 \\  
            & {\hspace{1em}}CL-MoE & \underline{90.74} & \underline{89.92} & \underline{89.43} & \underline{94.96} & 76.64 & 76.67 & 59.76 & 66.88 & 80.63 \\
            & {\hspace{1em}}HiDe-LLaVA & 83.67 & 84.06 & 82.48 & 91.44 & \underline{89.18} & 74.96 & 56.47 & 61.04 & 77.91 \\  
            & {\hspace{1em}}\textbf{CL-Anomaly} & \textbf{92.59} & \textbf{91.59} & \textbf{92.20} & \textbf{95.63} & \textbf{89.44} & \textbf{81.30} & \textbf{74.30} & \underline{73.64} & \textbf{86.34} \\  
            \midrule
            \multirow{5}{*}{\rotatebox{90}{\textbf{Last}}} 
            &{\hspace{1em}}LoRA & 81.58 & 82.36 & 80.28 & 92.43 & 60.94 & \underline{79.00} & \underline{70.92} & \textbf{76.36} & 77.98 \\
            & {\hspace{1em}}MoELoRA  & 86.98 & 85.96 & 85.93 & 93.40 & 65.41 & 74.68 & 58.57 & 71.82 & 77.84 \\  
            & {\hspace{1em}}CL-MoE & \underline{88.60} & \underline{87.95} & \underline{87.49} & \underline{93.73} & 69.29 & 73.73 & 56.97 & 66.88 & \underline{78.08} \\
            & {\hspace{1em}}HiDe-LLaVA & 75.31 & 76.39 & 73.47 & 89.13 & \textbf{87.88} & 68.54 & 54.78 & 61.04 & 73.32 \\  
            & {\hspace{1em}}\textbf{CL-Anomaly} & \textbf{92.06} & \textbf{90.37} & \textbf{91.75} & \textbf{94.05} & \textbf{87.88} & \textbf{81.24} & \textbf{73.90} & \underline{73.64} & \textbf{85.61} \\  
            \midrule
            
            \multirow{5}{*}{\rotatebox{90}{\textbf{BWT}}} 
            &{\hspace{1em}}LoRA & -5.13 & -5.59 & -8.76 & -4.42 & -24.36 & \underline{-3.02} & -3.98 & - & -7.89 \\   
            & {\hspace{1em}}MoELoRA & -2.72 & -2.22 & -4.23 & -3.08 & -22.43 & -4.62 & -8.16 & - & -6.78 \\  
            & {\hspace{1em}}CL-MoE & \underline{-1.87} & \underline{-1.90} & \underline{-3.48} & \underline{-2.99} & -19.53 & -5.96 & -5.58 & - & \underline{-5.90} \\
            & {\hspace{1em}}HiDe-LLaVA & -6.62 & -7.36 & -10.65 & -5.22 & \underline{-4.55} & -6.70 & \underline{-3.39} & - & -6.36 \\  
            & {\hspace{1em}}\textbf{CL-Anomaly} & \textbf{-0.07} & \textbf{-0.53} & \textbf{-0.79} & \textbf{-1.81} & \textbf{-2.95} & \textbf{0.21} & \textbf{-0.80} & - & \textbf{-0.96} \\  
        \bottomrule
    \end{tabular}}
    \label{tab:main}
\end{table*}

\subsection{Comparison with the State-of-the-art}
Table~\ref{tab:main} presents a comparison of our method with the baselines. Based on these results, we provide the following key findings:

\begin{itemize}
    \item CL-Anomaly exhibits outstanding overall performance, outperforming the current state-of-the-art. Compared with CL-MoE, the second-best method, it improves Avg, Last, and BWT by $5.68\%$, $7.44\%$, and $4.94\%$, respectively.
    Notably, CL-Anomaly approaches the upper bound of Avg and Last metrics in multi-task settings, demonstrating both its effectiveness and robust resistance to forgetting.

    \item Parameter sharing and isolation strategies exhibit advantages in different scenarios. Sharing-based methods (e.g., MoELoRA, CL-MoE) achieve strong performance on class-incremental tasks (WebAD-1 to WebAD-3) by effectively transferring knowledge among similar tasks. However, their transferability is limited in cross-domain increments, such as the medical BMAD dataset, due to substantial domain differences. Conversely, isolation-based methods like HiDe-LLaVA achieve modest performance on class-incremental tasks but remain stable in cross-domain scenarios, highlighting the rationale of CL-Anomaly’s isolation–sharing collaborative framework. Additional analyses of the baselines are provided in Appendix~\ref{apd:c3}.

    \item Most baseline methods perform poorly on 3D datasets due to large representation gaps between 2D and 3D tasks, causing severe semantic shifts. HiDe-LLaVA is constrained by its inference mechanism, which integrates non-top-level parameters across domains and modalities, causing knowledge entanglement and limiting performance. In contrast, CL-Anomaly leverages fully isolated private expert paths and a layer-adaptive knowledge transfer strategy to preserve the knowledge of each modality task and mitigate the impact of modality differences.
\end{itemize}

Given the inherent class imbalance in anomaly detection datasets and to ensure a fair comparison with existing continual learning approaches, we further report the $Accuracy$, $Precision$, $Recall$, and $F1\mathchar`\-score$ obtained after completing all training stages. As shown in Table~\ref{tab:ad}, CL-Anomaly achieves superior performance in all metrics, demonstrating the strong practicality of our method in anomaly detection tasks. More details and complete metrics for each task can be found in Appendix~\ref{apd:d1}.

\begin{table}[t]
    \centering
    \begin{minipage}[t]{0.47\textwidth}
        \centering
        \caption{Comprehensive metrics comparison in the last stage.}
        \label{tab:ad}
        \renewcommand{\arraystretch}{1.13}
        \resizebox{1\columnwidth}{!}{
        \begin{tabular}{l cccc | c}
            \toprule
            Method & Accuracy & Precision & Recall & F1-score & Average  \\
            \midrule
            Zero-shot & 73.99 & 80.46 & 65.39 & 69.52 & 72.34  \\
            Multi-task  & 88.38 &  90.13 & 83.83 & 86.18 & 87.13 \\
            \midrule
            LoRA & 77.98 & \underline{87.65} & 66.16 & \underline{74.74} & \underline{76.63}  \\
            MoELoRA  & 77.84 & 87.41 & 64.84 & 72.95 & 75.76  \\
            CL-MoE & \underline{78.08} & 85.51 & 66.30 & 73.47 & 75.84  \\
            HiDe-LLaVA & 73.32 & 71.14 & \underline{67.75} & 68.61 & 70.21  \\
            CL-Anomaly & \textbf{85.61} & \textbf{89.61} & \textbf{78.00} & \textbf{82.51} & \textbf{83.93}  \\
            \bottomrule
        \end{tabular}}
    \end{minipage}\hfill
    \begin{minipage}[t]{0.48\textwidth}
        \centering
        \caption{Ablation study of proposed CL-Anomaly.}
        \label{tab:ablation}
        \renewcommand{\arraystretch}{1.13}
        \resizebox{1\columnwidth}{!}{
        \begin{tabular}{lccc}
            \toprule
            Method & Avg & Last & BWT  \\
            \midrule
            Full Model & 86.34 & 85.61 & -0.96   \\
            \midrule
            w/o. PrivLoRA & 83.58 & 81.60 & -3.70  \\
            w/o. SharedLoRA & 84.97 & 82.63 & -3.29  \\
            w/o. Layer-Adaptive & 85.09 & 84.06 & -1.38   \\
            \bottomrule
        \end{tabular}}
    \end{minipage}
\end{table}

\subsection{Ablation Study}
We conducted a series of ablation experiments to verify the effectiveness of each component in CL-Anomaly, where Layer-Adaptive denotes the Layer-wise Adaptive Knowledge Transfer. As shown in the Table~\ref{tab:ablation}, each component contributes to the overall performance gain. Removing PrivLoRA leads to a significant performance drop, highlighting the importance of task-specific parameter independence for continual learning in anomaly detection. Disabling the shared path (SharedLoRA) restricts knowledge flow among similar tasks and results in notable degradation when the visual-anchor-based routing of PrivLoRA is suboptimal. The layer-adaptive knowledge transfer mechanism further promotes knowledge sharing across tasks while preserving layer-wise key information, effectively mitigating catastrophic forgetting.

\begin{table}[t]
\centering
\caption{Quantitative comparison of anomaly reasoning capabilities.}
\resizebox{\linewidth}{!}{
\begin{tabular}{l ccc ccc ccc}
\toprule
\multirow{2}{*}{Method}
& \multicolumn{3}{c}{Mvtec-AD}
& \multicolumn{3}{c}{BMAD}
& \multicolumn{3}{c}{Real3D-AD}
\\
\cmidrule(lr){2-4} \cmidrule(lr){5-7} \cmidrule(lr){8-10} 
& ROUGE-L & SBERT & LLM-Score & ROUGE-L & SBERT & LLM-Score & ROUGE-L & SBERT & LLM-Score \\
\midrule
Zero-shot & 41.00 & 79.81 & 6.50 & 31.73 & 68.67 & 5.02 & 34.38 & 73.53  & 4.67 \\
CL-MoE & 43.47 & 83.81 & 7.17 & 38.24 & 73.25 & 6.29 & 41.68 & 82.03 & 5.94 \\
CL-Anomaly & \textbf{44.65} & \textbf{85.40} & \textbf{7.64} & \textbf{40.70} & \textbf{76.94} & \textbf{6.77} & \textbf{43.14} & \textbf{83.27} & \textbf{6.60} \\
\bottomrule
\end{tabular}}
\label{tab:exp-reasoning}
\end{table}

\subsection{Further Analysis}


\noindent\textbf{Evaluation on Anomaly Reasoning.}
To further validate the performance of our model in anomaly reasoning under continual learning, we evaluated CL-Anomaly and CL-MoE on three representative datasets: the industrial domain (Mvtec-AD), the medical domain (BMAD), and the 3D modality (Real3D-AD), reporting results after all tasks were learned. We evaluate reasoning performance using Rouge-L, SBERT, and LLM-Score metrics. Specifically, Rouge-L evaluates the structural and lexical consistency between model outputs and reference answers; SBERT evaluates the semantic similarity under varied linguistic formulations; and LLM-Score leverages a large language model as an automatic judge to compare candidate and reference responses, providing a comprehensive measure of correctness, completeness, and reasoning quality. As shown in Table~\ref{tab:exp-reasoning}, our method outperforms the current continual learning SOTA, CL-MoE, in terms of expression fidelity, semantic consistency, and holistic reasoning quality, demonstrating that CL-Anomaly preserves robust anomaly reasoning capabilities for previously learned tasks even after multi-stage training. More details are provided in Appendix~\ref{apd:d2}.

\begin{figure}[t]
    \centering
    \begin{minipage}[t]{0.48\textwidth}
        \centering
        \begin{subfigure}[b]{0.48\textwidth}
            \centering
            \includegraphics[width=\linewidth]{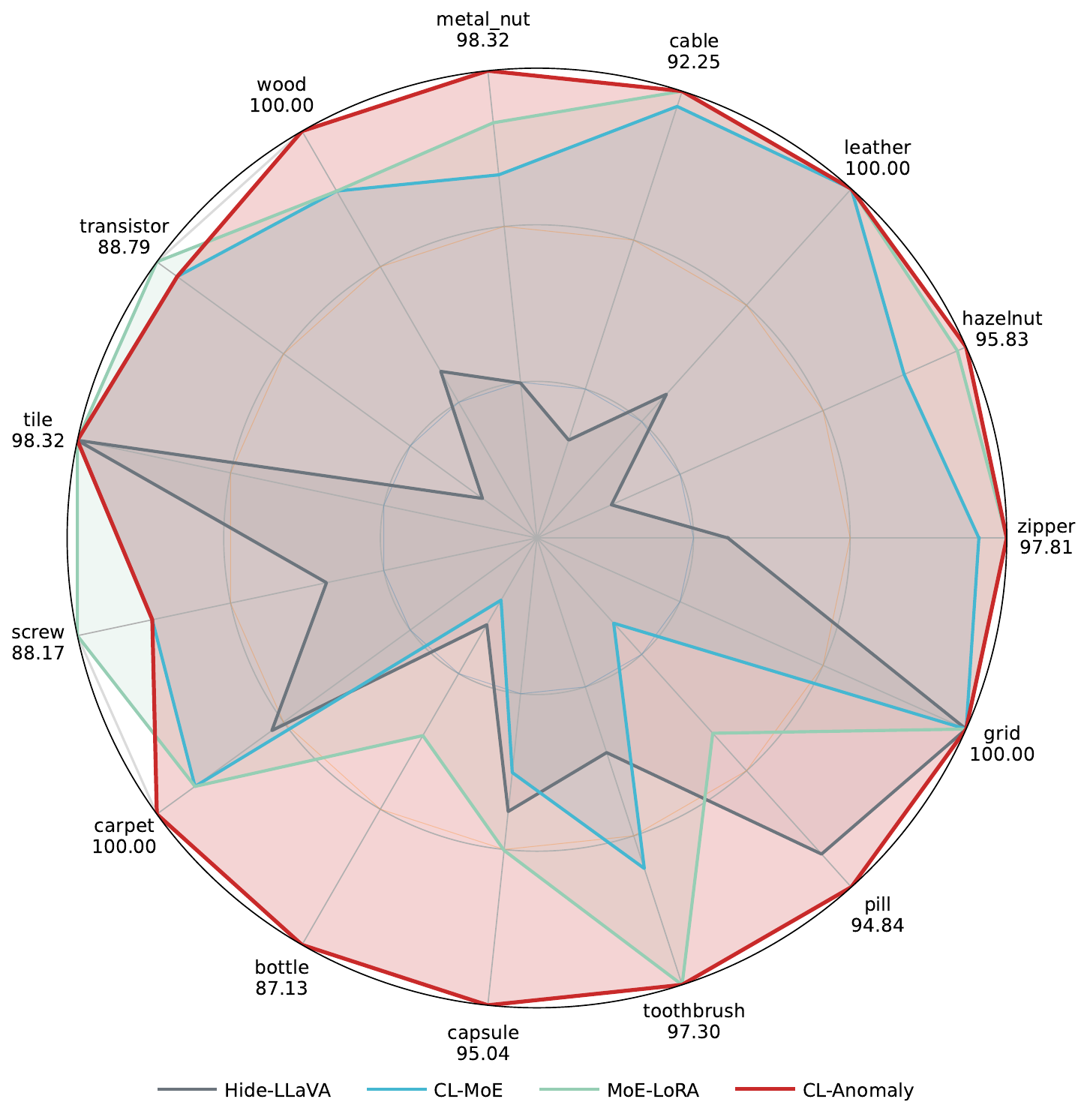}
            \caption{Accuracy}
            \label{fig:radar_acc}
        \end{subfigure}%
        \hfill
        \begin{subfigure}[b]{0.48\textwidth}
            \centering
            \includegraphics[width=\linewidth]{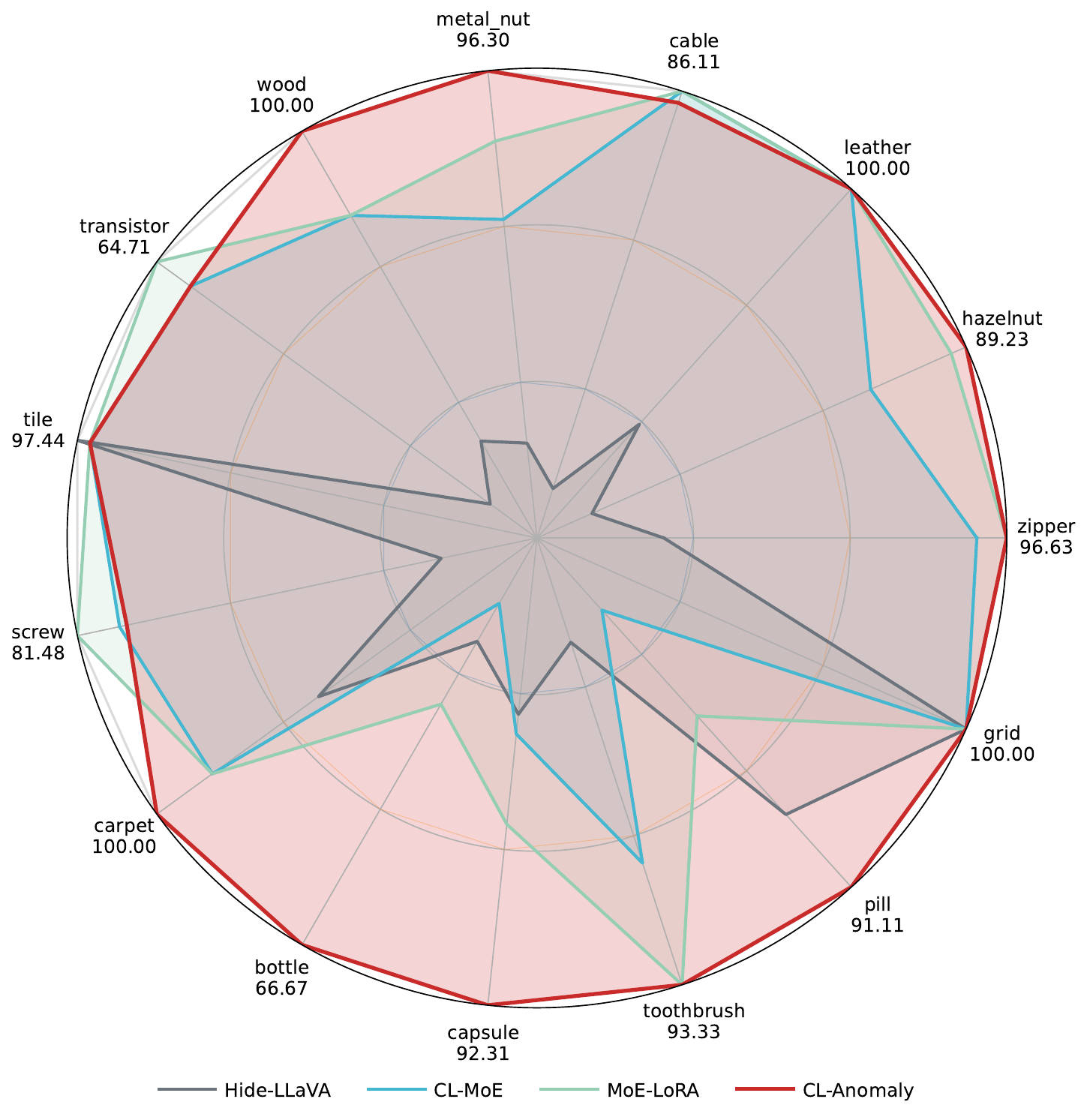}
            \caption{F1-score}
            \label{fig:radar_f1}
        \end{subfigure}
        \caption{Visualization of CL-Anomaly, and the SOTA continual learning approaches under class-incremental learning on practical single-anomaly scenarios.}
        \label{fig:radar}
    \end{minipage}
    \hfill
    \begin{minipage}[t]{0.48\textwidth}
        \centering
        \begin{subfigure}[b]{0.48\textwidth}
            \centering
            \includegraphics[width=\linewidth]{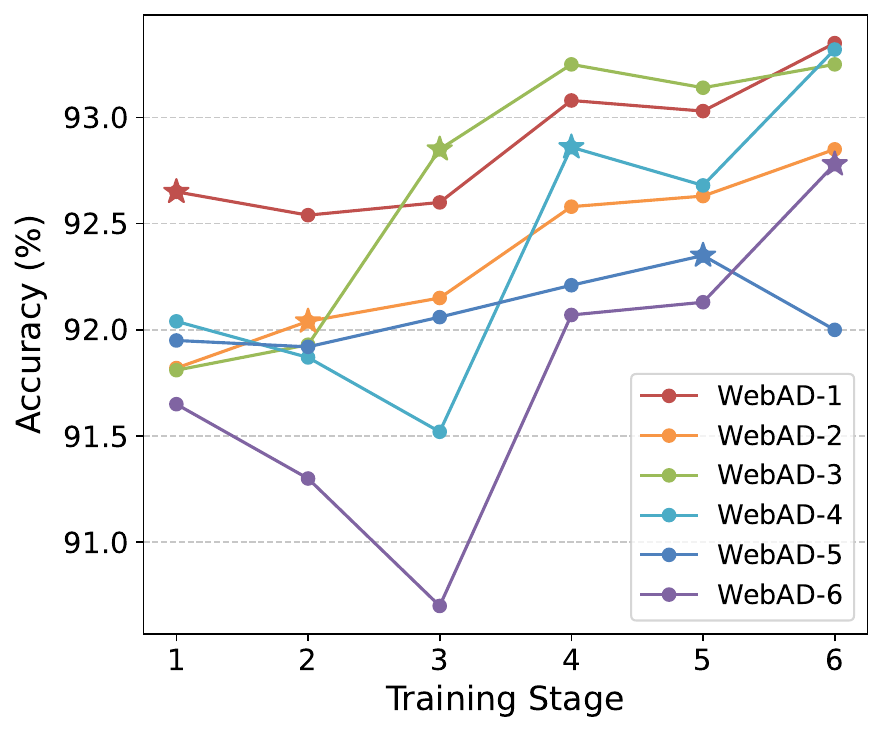}
            \caption{}
            \label{fig:cil_acc}
        \end{subfigure}%
        \hfill
        \begin{subfigure}[b]{0.48\textwidth}
            \centering
            \includegraphics[width=\linewidth]{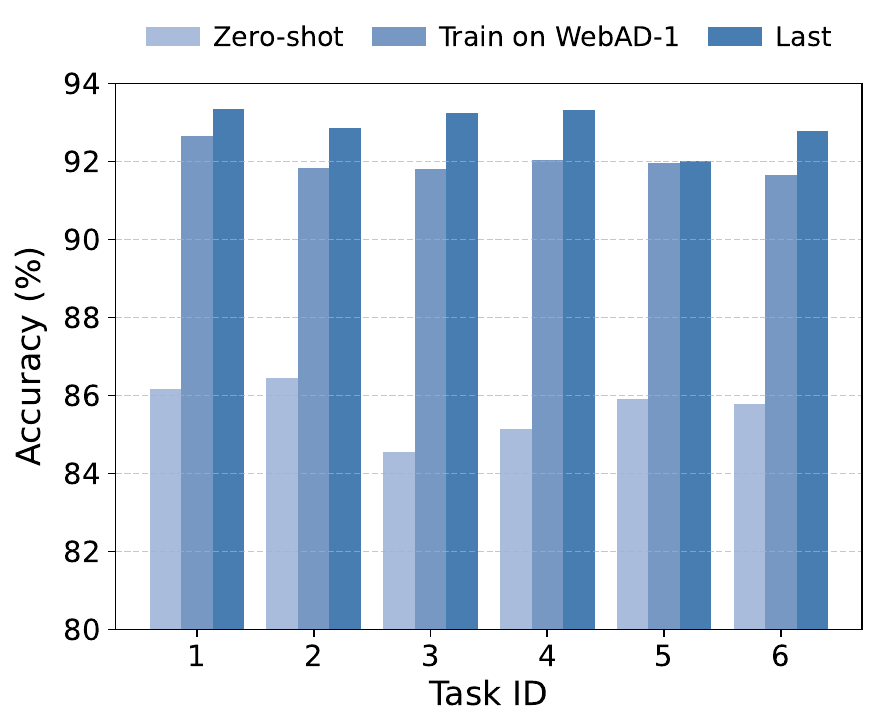}
            \caption{}
            \label{fig:class_acc}
        \end{subfigure}
        \caption{(a) Accuracy of each task across stages in CIL, $\star$ indicates the target task being learned at that stage. (b) Accuracy under CIL compared to zero-shot.}
        \label{fig:cil_analysis}
    \end{minipage}
\end{figure}

\noindent\textbf{Experiments in Traditional Continuous AD Scenarios.}
Tables~\ref{tab:main} and~\ref{tab:ad} demonstrate the continual anomaly detection capability of our method across multiple complex scenarios. We further validate CL-Anomaly's continual learning performance under single-anomaly scenarios, which are prevalent in practical continual AD systems. Following the experimental setting of traditional continual anomaly detection works\cite{tang2024incremental,liu2024unsupervised}, we partition the Mvtec-AD dataset into 15 category-based subsets and learn each subtask sequentially. We evaluate the model performance after training on all subtasks. As shown in \cref{fig:radar}, CL-Anomaly maintains superior detection capability despite multi-stage training on non-target category tasks, demonstrating the effectiveness of our method in practical continual anomaly detection systems. Detailed metrics and further descriptions are provided in Appendix~\ref{apd:d3}.

\noindent\textbf{Analysis of Knowledge Transfer in CIL.}
We further introduce additional data categories from the WebAD dataset to construct a larger-scale class incremental learning (CIL) scenario, which is the most common task setting in real-world anomaly detection. Unlike the main experiments, this study reports task accuracy at each stage, regardless of whether the model has encountered the corresponding task. As shown in \cref{fig:cil_acc}, task performance generally improves throughout continual learning, even after the task has been fully trained, indicating that CL-Anomaly enables positive backward knowledge transfer, where learning new tasks boosts the performance of previously learned ones. Moreover, we compare task accuracy under three conditions: zero-shot, after training only on WebAD-1, and after completing all tasks, as shown in \cref{fig:class_acc}. After training on only the first task, CL-Anomaly markedly surpasses the zero-shot baseline across all tasks, demonstrating strong forward knowledge transfer, where learning early tasks improves performance on subsequently unseen tasks.
\begin{figure*}[t]
\centering

\begin{minipage}[t]{0.52\textwidth}
 \centering
    
    \adjustbox{valign=t}
    {\includegraphics[width=\linewidth]{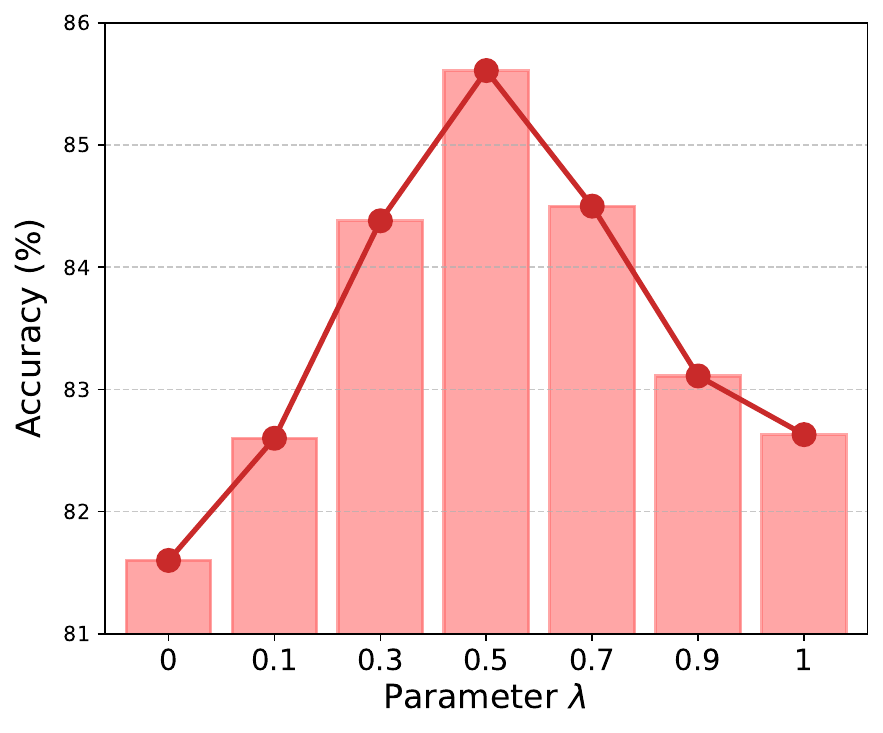}}
    \captionof{figure}{Average last accuracy of $\lambda$.}
    \label{fig:lambda}
    
    \adjustbox{valign=t}
    {\includegraphics[width=\linewidth]{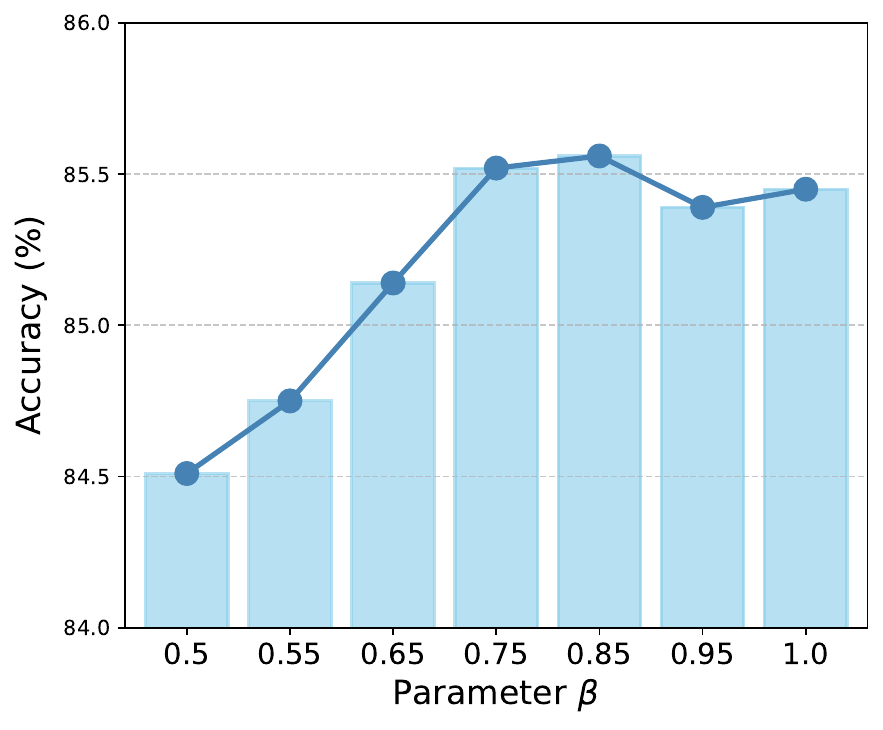}}
    \captionof{figure}{Average last accuracy of $\beta$.}
    \label{fig:beta}
    
\end{minipage}
\hfill
\begin{minipage}[t]{0.46\textwidth}
 \centering
    
    \captionof{table}{Zero-shot results. \textit{multi} denotes multi-task joint training.}
    \label{tab:zsad}
    \resizebox{\linewidth}{!}{
    \begin{tabular}[t]{@{}l |cc@{}}
        \toprule
        Method & VisA & DTD \\
        \midrule
        IAD-R1-7B (LLaVA) & 76.09 & 91.18 \\
        IAD-R1-7B (Qwen) & 65.47 & 77.91 \\
        AnomalyOV-7B & 78.53 & 94.33 \\
        \midrule
        Ours-7B & 75.41 & 95.86 \\
        Ours-multi-7B & 76.43 & 96.63 \\
        \bottomrule
    \end{tabular}}
    
    \captionof{table}{Analysis on the number of shared experts.}
    \label{tab:num}
    \resizebox{\linewidth}{!}{
    \begin{tabular}[t]{@{}ccccc@{}}
        \toprule
        Method  & $\langle n, k \rangle$  & Avg & Last & BWT  \\
        \midrule
        \multirow{4}{*}{CL-Anomaly}  
        & $\langle 1, 1 \rangle$ & 85.30 & 82.29 & -3.68 \\
        & $\langle 2, 1 \rangle$ & 86.60 & 85.43 & -1.68 \\
        & $\langle 4, 1 \rangle$ & 86.31 & 85.52 & \textbf{-0.96} \\
        & $\langle 8, 2 \rangle$ & \textbf{86.75} & \textbf{85.87} & -1.15 \\
        \bottomrule
    \end{tabular}}
    
    \captionof{table}{Performance under different task orders.}
    \label{tab:order}
    \resizebox{\linewidth}{!}{
    \begin{tabular}[t]{@{}l ccc ccc@{}}
        \toprule
        \multirow{2}{*}{Method}
        & \multicolumn{3}{c}{Forward}
        & \multicolumn{3}{c}{Random} \\
        \cmidrule(lr){2-4} \cmidrule(lr){5-7}
        & Avg & Last & BWT & Avg & Last & BWT \\
        \midrule
        CL-MoE & 80.63 & 78.08 & -5.90 & 81.18 & 77.99 & -5.70 \\
        CL-Anomaly  & 86.34 & 85.61 & -0.96 & 84.81 & 84.84 & -0.72 \\
        \bottomrule
    \end{tabular}}

\end{minipage}

\end{figure*}

\noindent\textbf{Performance on ZSAD.} 
We further compare CL-Anomaly with powerful zero-shot anomaly detection MLLMs\cite{xu2025towards,li2026iad} under the zero-shot setting. Specifically, we evaluate on the DTD and VisA subsets (1,000 anomalous and 761 normal images). The results are presented in Table~\ref{tab:zsad}. Despite not being explicitly designed for zero-shot generalization, CL-Anomaly achieves competitive performance against these specialized ZSAD methods, highlighting its strong inherent transferability.

\noindent\textbf{Impact of Hyperparameter $\lambda$.} \cref{fig:lambda} illustrates the impact of the hyperparameter $\lambda$  in Eq.2. Both excessively small and large values of $\lambda$  degrade performance, while $\lambda=0.5$  strikes an optimal balance, effectively mediating the trade-off between shared and private experts.

\noindent\textbf{Impact of Hyperparameter $\beta$.} \cref{fig:beta} shows the impact of hyperparameter $\beta$ in Eq.9. Larger $\beta$ strengthens key knowledge retention and fades irrelevant knowledge; moderate increases improve performance, yet $\beta$ near 0.5 degenerates into simple fusion and performance declines, demonstrating LAKT's robustness for $\beta$ within a reasonable range. And considering the relative balance between prior and novel knowledge, we set the default $\beta_{l,e}$ to 0.5.

\noindent\textbf{Impact of Shared Expert Number.}
Table~\ref{tab:num} presents the impact of the number of shared experts $n$ and key experts per layer $k$ on model performance. CL-Anomaly maintains stable performance across a reasonable parameter range, and even with only two shared experts, it achieves competitive results, enabling efficient cross-task knowledge transfer with minimal additional shared experts.

\noindent\textbf{Impact of Task Order. }
We further investigate the impact of different task orders on CL-Anomaly performance. Specifically, we randomly generated the following task order: WebAD-3 $\rightarrow$ Real3D-AD $\rightarrow$ WebAD-1 $\rightarrow$ Anomaly-Shapenet $\rightarrow$ MvTec-AD $\rightarrow$ WebAD-2 $\rightarrow$ MvTec-3D $\rightarrow$ BMAD, and compared the results with CL-MoE, as shown in the Table~\ref{tab:order}. Although the order of knowledge transfer across tasks inevitably affects performance, CL-Anomaly exhibits relatively stable performance and outperforms the state-of-the-art method CL-MoE.

\begin{figure}[t]
    \centering
    \begin{minipage}[t]{0.52\textwidth}
        \centering
        \includegraphics[width=\linewidth]{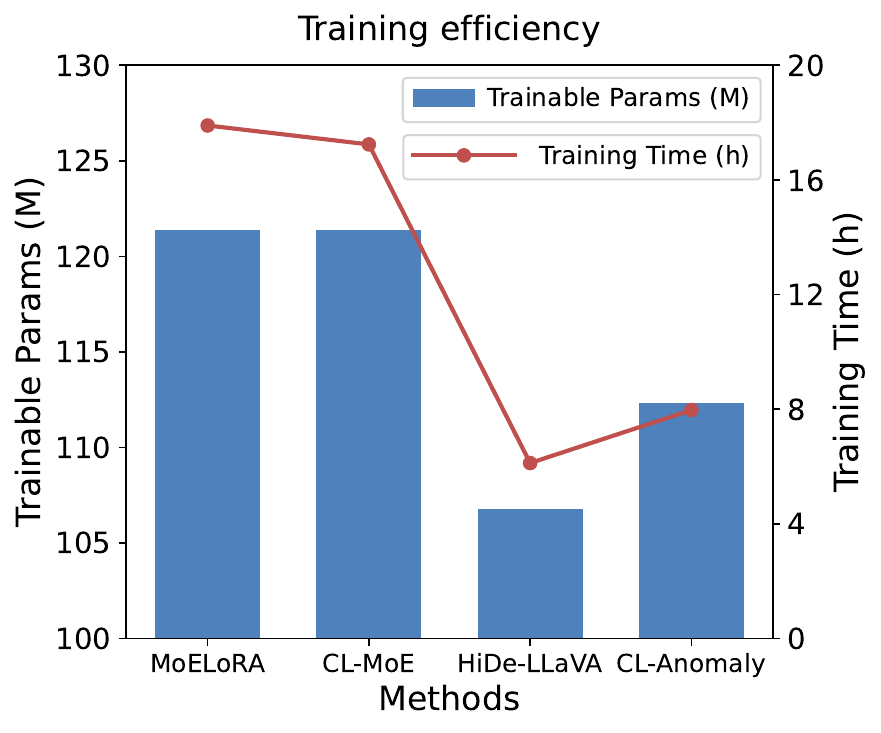}
        \caption{Training efficiency comparison of our method and representative approaches.}
        \label{fig:time}
    \end{minipage}\hfill
    \begin{minipage}[t]{0.45\textwidth}
        \centering
        \includegraphics[width=0.9\linewidth]{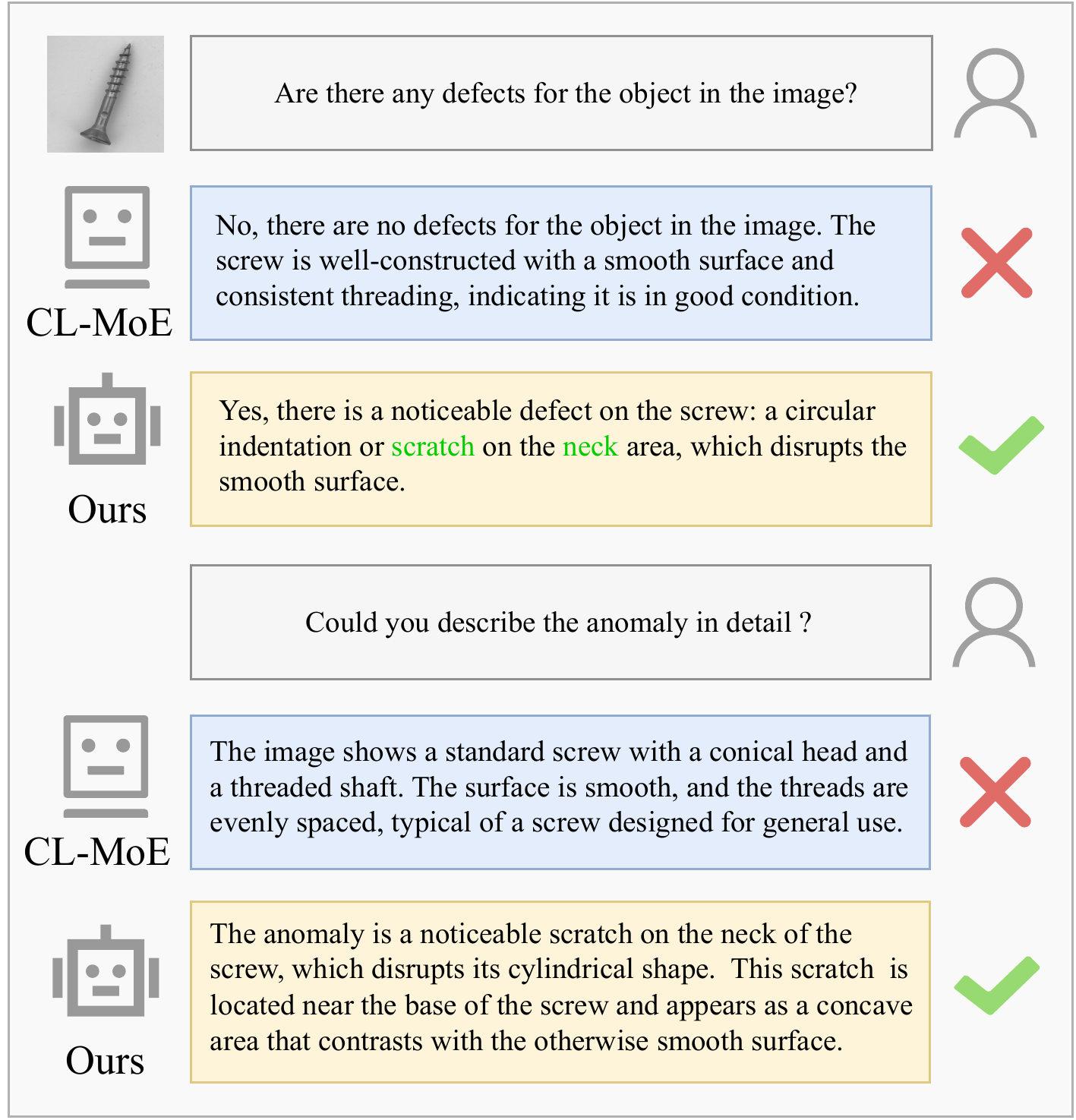}
        \caption{Visualization of proposed CL-Anomaly and CL-MoE.}
        \label{fig:example}
    \end{minipage}
\end{figure}

\noindent\textbf{Analysis of Training Parameters and Time.}
\cref{fig:time} presents the trainable parameters and training time of CL-Anomaly compared with baseline methods. MoELoRA and CL-MoE incur the highest training cost, while CL-Anomaly achieves comparable efficiency to HiDe-LLaVA with markedly improved performance. Compared with the pretrained backbone ($8020.92$ M), CL-Anomaly incurs only a $1.4\%$ parameter overhead, demonstrating superior parameter efficiency and deployment feasibility.

\noindent\textbf{Example Analysis.}
\cref{fig:example} shows an example from industrial scenario. CL-Anomaly accurately detects anomalies while providing clear visual evidence and precise descriptions of anomaly regions, demonstrating strong anomaly understanding and reasoning capabilities. Meanwhile, the support for multi-round human–machine interaction makes it more user-friendly than traditional detection models. More examples are provided in Appendix~\ref{apd:vis}.

%% file: sec/7_acknowledgements.tex
\section*{Acknowledgements}
This work was supported by the National Natural Science Foundation of China (No. 62476124, 62276134), the Fundamental and Interdisciplinary Disciplines Breakthrough Plan of the Ministry of Education of China (JYB2025XDXM902), the Natural Science Foundation of Jiangsu Province (No. BK20242015), the Gusu Innovation and Entrepreneur Leading Talents (No. ZXL2025322), and Nanjing University-China Mobile Communications Group Co.,Ltd. Joint Institute.

%% file: Supplementary_Material/appendix.tex
\newpage
\appendix
\begin{center}
	\large \textbf{Appendix for CL-Anomaly: Layer-Adaptive Mixture-of-Experts with Multimodal Large Language Model for Continual Learning in Anomaly Detection}
\end{center}
\par
This supplementary appendix contains the following  six parts: \textcolor{blue}{\cref{apd:mot_pro} }Motivation and Problem Definition \textcolor{blue}{\cref{apd:dataset}} Details of the dataset \textcolor{blue}{\cref{apd:imp}} Implementation Details \textcolor{blue}{\cref{apd:exp}} Experimental Details and Additional Results \textcolor{blue}{\cref{apd:lim}} Limitation \textcolor{blue}{\cref{apd:vis}} Visualization Examples

\section{Motivation and Problem Definition}
\label{apd:mot_pro}
\subsection{Motivation and Practicality. }
We elaborate on the motivation and practicality of our work from two perspectives: (1) the advantages of MLLMs in anomaly detection, and (2) the importance of the continual learning paradigm in anomaly detection.

\noindent\textbf{Advantages of MLLMs in Anomaly Detection. }
(1) Traditional non-MLLM anomaly detection methods are typically confined to a single domain or specific modality, making effective transfer to broader application scenarios challenging. In contrast, MLLMs inherently possess cross-modal and cross-domain processing capabilities, enabling a single unified model to adapt to diverse scenarios. (2) Conventional methods typically output continuous anomaly scores, while some approaches can produce pixel-level anomaly localization maps; however, both lack in-depth, convincing explanations for the rationale behind anomaly judgments.  Moreover, practical deployment often relies on tedious threshold tuning. In comparison, MLLMs can directly output qualitative anomaly decisions with verifiable reasoning, not only providing explicit determinations of whether a target is anomaly but also offering semantic-level interpretations of anomalous regions and types, which is more user-friendly and actionable in practice. (3) The definition of anomalies is highly context-dependent, varying with specific application scenarios and task objectives. Benefiting from their strong human–machine interaction and instruction-following capabilities, MLLMs enable users to dynamically adjust anomaly decision boundaries according to current scene requirements. Furthermore, through multi-turn dialogue, users can further examine detected instances, exploring detailed anomaly characteristics and related issues such as cause analysis and prevention strategies. 

\noindent\textbf{Importance of Continual Learning in Anomaly Detection. }An important motivation stems from the practical characteristics of real-world anomaly detection systems, valuable data typically accumulates progressively, while retraining models from scratch by jointly integrating newly collected and historical data would incur substantial computational overhead. For example, in medical imaging diagnosis systems, valuable data expands dynamically with increasing patient cases. Therefore, it is crucial for models to acquire new knowledge from incoming data while retaining previously learned information, especially in scenarios where historical data may be unavailable due to privacy constraints.

In addition, detection targets themselves often evolve. For instance, in manufacturing factories undergoing business expansion, the variety of product lines often increases continuously, requiring models to accommodate detection demands for both existing and newly introduced product categories. Such scenarios correspond to the classic class-incremental learning paradigm in anomaly detection, which we validate in Section 5.4 (\textit{Experiments in Traditional Continuous AD Scenarios}) and Appendix~\ref{apd:d3}.

More generally, when the objectives or anomaly decision criteria of detection tasks shift significantly over time, successive tasks may exhibit heterogeneous modalities or conflicting domain distributions, leading to a more challenging multi-scenario continual anomaly detection setting. We adopt this setting as our main experimental setup to fully evaluate the stability and adaptability of the proposed method in general continual anomaly detection tasks.

\begin{figure}[t]
\centering
  \includegraphics[width=\linewidth ]
  {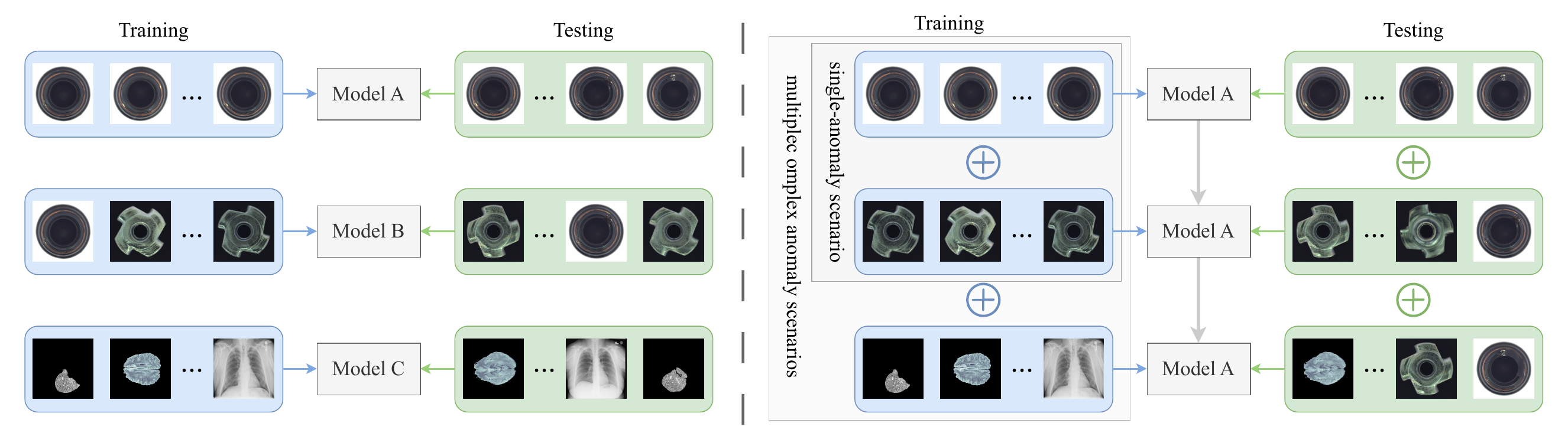}
  \caption{(Left) Many existing methods cannot dynamically update knowledge, often requiring the deployment of multiple specialized models to address evolving detection scenarios. (Right) Our method enables a single model to continually learn both intra-task and cross-task anomaly knowledge, accommodating broader application scenarios.}
  \label{apd:mot}
\end{figure}

Many existing anomaly detection paradigms, once trained, cannot dynamically update knowledge, as illustrated in \cref{apd:mot} (left), limiting applicability in continually evolving real-world detection scenarios. In contrast, as shown in \cref{apd:mot} (right), our method not only supports intra-task continual detection in single-anomaly scenarios but also stably updates knowledge across complex multi-anomaly scenarios.

\subsection{About Problem Definition. }
Our work focuses on adapting the MLLM-based continual learning paradigm to anomaly detection tasks. During the continual learning process, the model learns tasks sequentially, progressing stage by stage, aiming to effectively acquire new knowledge while preventing forgetting of previously learned tasks. 

The definition of each stage is flexible: a stage can correspond either to a complete dataset within a specific scenario that contains multiple categories, or to a single category within one scenario. In our main experiments, we define each stage as a full dataset containing multiple fine-grained categories. These datasets span diverse domains and modalities, enabling a comprehensive evaluation of the model’s stability in complex multi-scenario continual anomaly detection. 

Aligning with prior continual anomaly detection works\cite{tang2024incremental,liu2024unsupervised} that focus on class-incremental learning within a single scenario, where each stage comprises only a single category from the dataset, we also discuss this less challenging yet practically encountered setting in realistic anomaly detection systems in Section 5.4 (\textit{Experiments in Traditional Continuous AD Scenarios}) and Appendix~\ref{apd:d3}.

\section{Details of the dataset}
\phantomsection\label{apd:dataset}
We used all sub-datasets contained in Anomaly-Instruct-125K as sequential tasks for continual learning. Each task is randomly partitioned into training and test splits using an 8:2 ratio. As detailed in Table~\ref{apd-tab:dataset}, we provide the domain and modality of each dataset together with the number of sample classes, and the corresponding sample counts used during training and testing. For 3D data, we follow the processing protocol of Anomaly-Instruct-125K by converting each 3D point cloud into a set of multi-view images.

Due to the large scale and high class diversity of WebAD, and considering the computational cost of using the full data, we randomly select several subsets for our main experiments to evaluate class-incremental learning capability. For the other datasets, we use the full amount of data. The specific WebAD class subsets used in our experiments are listed in Table~\ref{apd-tab:webad}. 
\begin{figure}[t]
    \centering
    \begin{minipage}[t]{0.44\textwidth}
        \centering
        \captionof{table}{Detailed information of each dataset.}
        \resizebox{0.96\linewidth}{!}{
        \begin{tabular}{lccccc}
        \toprule
            Dataset & Domain & Modality & $|\mathrm{Class}|$ & Train & Test  \\
            \midrule
            WebAD-1 & Industrial & 2D  & 35 & 7404 & 1851 \\
            WebAD-2 & Industrial & 2D & 35 & 7434 & 1859  \\
            WebAD-3 & Industrial & 2D & 35 & 6935 & 1734  \\
            Mvtec-AD & Industrial & 2D & 15 & 7393 & 1849 \\
            BMAD & Medical  & 2D & 6 & 3400 & 850  \\
            Mvtec-3D & Industrial & 3D & 10 & 4627 & 1157 \\
            Real3D-AD  & Industrial & 3D & 12 & 2006 & 502  \\
            Anomaly-Shapenet & Industrial & 3D  & 52 & 3078 & 770\\
        \bottomrule
        \end{tabular}}
        \label{apd-tab:dataset}
    \end{minipage}\hfill
    \begin{minipage}[t]{0.56\textwidth}
        \centering
        \captionof{table}{Sample class composition of the WebAD subset used.}
        \resizebox{\linewidth}{!}{
        \begin{tabular}{lc}
        \toprule
            Dataset & Class  \\
        \midrule
            WebAD-1 & \makecell[l]{
            $Bunsen\_burner, banana, bed, bookcase, broom, camera\_tripod, cell\_phone, coffee\_maker, desk$ \\
            $drone, escalator, fire\_hydrant, freezer, glue, hard\_hat, hose, karaoke\_machine, lawn\_mower$ \\
            $mailbox, mirror, outlet, person, police\_car, recycling\_bin, safety\_vest, security\_camera, sink$ \\
            $soap\_dispenser, stethoscope, sunglasses, tape\_dispenser, thermos, trailer, vacuum\_cleaner, wheelchair$
            } \\
        \midrule
            WebAD-2 & \makecell[l]{
            $IV\_stand, bandage, belt, bookshelf, bucket, camping\_stove, chair, compass, dining\_table$ \\
            $drum\_set, excavator, fire\_truck, frisbee, goggles, hat, hospital\_bed, kettle, leaf\_blower$ \\
            $map, mittens, oven, petri\_dish, pot, refrigerator, sandals, shelf, skateboard$ \\
            $soccer\_ball, stool, surfboard, tape\_measure, thermostat, train, vase, whistle$
            } \\
        \midrule
            WebAD-3 & \makecell[l]{
            $abacus, barbell, bench, boot, bulldozer, camping\_tent, chalkboard, computer, dishwasher$ \\
            $dryer, exercise\_bike, first\_aid\_kit, furniture, golf\_club, headphones, hot\_dog, kettlebell, level$ \\
            $marker, modem, package, phone, potted\_plant, remote, sandwich, ship, ski$ \\
            $sofa, stop\_sign, surgical\_tools, taxi, tie, trash\_can, vending\_machine, whiteboard$
            } \\
        \bottomrule
        \end{tabular}}
        \label{apd-tab:webad}
    \end{minipage}
\end{figure}

\section{Implementation Details}
\label{apd:imp}

\subsection{Evaluation Metrics}
\phantomsection\label{apd:c1}
\noindent\textbf{Metrics Definition. }
We evaluate continual learning performance using $Avg$, $Last$, and $BWT$ metrics. $Avg$ denotes the average accuracy throughout the training phase. Let $A_{t,i}$ represent the test performance on task $i$ after completing training on task $n$, then $Avg_{i}$ can be expressed as: $Avg_{i}=\frac{1}{T}\sum_{t=i}^{T}A_{t,i}$, where $T$ denotes the number of tasks. $Last$ represents the accuracy after training on all tasks: $Last_{i}=A_{T,i}$. $BWT$ measures the degradation of previously learned tasks after learning new tasks, defined as: $BWT_{i}=\frac{1}{T-i}\sum_{t=i+1}^{T}(A_{t,i}-A_{i,i})$. Considering the inherent class imbalance in anomaly detection, we additionally adopt $Precision$, $Recall$, and $F1\mathchar`\-score$ for a fairer comparison. We evaluate the model's performance on each task after completing the final training task. Consequently, $Precison_{i}=P_{T,i}$, $Recall_{i}=R_{T,i}$, $F1\mathchar`\-score_{i}=F_{T,i}$.

\noindent\textbf{Rationale for Metric Selection. }
$Avg$, $Last$, and $BWT$ are typically computed based on Accuracy and serve as standard metrics in continual learning to reflect the model's anti-forgetting capability. In anomaly detection, datasets commonly exhibit a certain degree of class imbalance, making $Accuracy$ alone insufficient to fairly reflect model performance. Accordingly, following the evaluation protocol adopted in prior MLLM-based anomaly detection methods~\cite{xu2025towards}, we additionally report $Precision$, $Recall$, and $F1\mathchar`\-score$. The comprehensive evaluation via these metrics enables a more reliable assessment of the model's practical anomaly detection capability.

\noindent\textbf{Why AUROC Is Not Adopted. }
Unlike conventional AD methods that output continuous anomaly scores, MLLMs produce discrete Yes/No decisions. Accordingly, $Accuracy$ rather than $AUROC$ is adopted by advanced MLLM-based AD methods, such as AnomalyR1\cite{chao2025anomalyr1}, IAD-R1\cite{li2025iad}, and AnomalyOV\cite{xu2025towards}(Notably, AnomalyOV comprises two components: anomaly expert and MLLM. The anomaly expert is trained in alignment with the zero-shot anomaly detection paradigm and outputs continuous anomaly scores, enabling Image-level $AUROC$ comparison with conventional ZSAD methods, see Table 2 in the original paper. Upon integrating the anomaly expert into the MLLM, the unified model outputs discrete Yes/No decisions. In this setting, performance is evaluated using classification metrics such as $Accuracy$ and $Precision$, as presented in Table 4 of the original paper). And due to this paradigm difference, these methods, including ours, are compared with MLLM methods rather than traditional AD approaches.

\subsection{Details of the comparison methods}
\phantomsection\label{apd:c2}
\noindent\textbf{LoRA. } Low-Rank Adaptation (LoRA) is a parameter-efficient model fine-tuning method that adapts a pretrained model to specific tasks by injecting low-rank update matrices into its weight parameters, without modifying the original weights. 

\noindent\textbf{MoELoRA. } This approach efficiently combines the Mixture-of-Experts (MoE) mechanism with LoRA for parameter-efficient fine-tuning. It adapts large language models to multiple tasks by activating specific LoRA experts through trainable routers.

\noindent\textbf{CL-MoE. } Building upon LoRA-based MoE, CL-MoE incorporates a Dual-Router MoE (RMoE) to simultaneously encode instance-level and task-level expert preferences, together with a Dynamic Momentum MoE (MMoE) that updates shared and task-specific experts through a momentum-based mechanism designed to preserve previously learned knowledge.

\noindent\textbf{HiDe-LLaVA. } This method introduces a hierarchical parameter-decoupling framework for continual instruction tuning of multimodal large language models. During training, it extracts visual and textual prototypes for each input via the vision and text encoders while updating the expert parameters of the current task. During inference, it selects the most suitable expert by measuring the similarity between the input’s visual features, textual query, and the stored prototypes, and employs a hybrid strategy that keeps top-layer experts independent while merging the parameters of lower-layer experts.

\subsection{Hyperparameter Settings for each method}
\phantomsection\label{apd:c3}

\noindent\textbf{LoRA. } This is the standard LoRA fine-tuning strategy, with the low-rank matrices set to a rank of 36 and the scaling factor $\alpha$ set to 72, consistent with our method.

\noindent\textbf{MoELoRA. } Using a vanilla LoRA-based MoE architecture, we set the number of experts to 12, matching the total count of our private and shared experts. Similarly, the rank of 36 with a scaling factor $\alpha$ of 72.

\noindent\textbf{CL-MoE. } Similar to MoELoRA, we set the number of experts to 12, the rank to 36, and the scaling factor $\alpha$ to 72. Notably, in the original work, the RMoE process used $K=2$ when the number of experts was 8; in our reproduction, we proportionally scale $K$ to 3. All other hyperparameters remain consistent with the original.

\noindent\textbf{HiDe-LLaVA. } We set the rank to 36 and the scaling factor $\alpha$ to 72, while all other hyperparameters follow the original settings.

\begin{table}[h]
    \centering
    \caption{Accuracy of each method across all tasks.}
    \label{ap-tab:acc}
    \resizebox{\linewidth}{!}{
    \begin{tabular}{l cccc cccc c}
        \toprule
        Method & WebAD-1 & WebAD-2 & WebAD-3 & Mvtec-AD & BMAD & Mvtec-3D & Real3D-AD & Anomaly-Shapenet & \textit{Average}  \\
        \midrule
        Zero-shot & 86.17 & 86.44 & 84.54 & 92.05 & 61.53 & 69.06 & 55.58 & 56.62 & 73.99 \\
        Multi-task  & 93.41 & 92.47 & 93.48 & 97.73 & 92.59 & 83.92 & 77.69 & 75.71 & 88.38  \\
        \midrule
        LoRA & 81.58 & 82.36 & 80.28 & 92.43 & 60.94 & \underline{79.00} & \underline{70.92} & \textbf{76.36} & 77.98 \\
        MoELoRA  & 86.98 & 85.96 & 85.93 & 93.40 & 65.41 & 74.68 & 58.57 & 71.82 & 77.84 \\
        CL-MoE & \underline{88.60} & \underline{87.95} & \underline{87.49} & \underline{93.73} & 69.29 & 73.73 & 56.97 & 66.88 & \underline{78.08} \\
        HiDe-LLaVA & 75.31 & 76.39 & 73.47 & 89.13 & \textbf{87.88} & 68.54 & 54.78 & 61.04 & 73.32 \\
        CL-Anomaly& \textbf{92.06} & \textbf{90.37} & \textbf{91.75} & \textbf{94.05} & \textbf{87.88} & \textbf{81.24} & \textbf{73.90} & \underline{73.64} & \textbf{85.61} \\
        \bottomrule
    \end{tabular}}
\end{table}

\begin{table}[h]
    \centering
    \caption{Precision of each method across all tasks.}
    \label{ap-tab:pre}
    \resizebox{\linewidth}{!}{
    \begin{tabular}{l cccc cccc c}
        \toprule
        Method & WebAD-1 & WebAD-2 & WebAD-3 & Mvtec-AD & BMAD & Mvtec-3D & Real3D-AD & Anomaly-Shapenet & \textit{Average}  \\
        \midrule
        Zero-shot & 97.59 & 96.30 & 97.33 & 82.35 & 81.85 & 52.19 & 55.56 & 80.49 & 80.46 \\
        Multi-task  & 91.19 & 89.38 & 91.51 & 95.98 & 94.69 & 77.59 & 90.50 & 90.20 & 90.13 \\
        \midrule
        LoRA & 97.08 & 96.13 & 97.15 & 88.74 & 91.10 & \underline{69.12} & 70.48 & \underline{91.39} & \underline{87.65} \\
        MoELoRA  & \underline{98.07} & \underline{96.68} & \textbf{98.25} & \underline{89.01} & \underline{92.09} & 59.30 & \textbf{76.34} & 89.57 & 87.41 \\
        CL-MoE & \textbf{98.15} & \textbf{96.73} & \underline{98.04} & 88.98 & \textbf{92.27} & 58.15 & 66.14 & 85.64 & 85.51 \\
        HiDe-LLaVA & 72.31 & 71.99 & 73.59 & 80.94 & 90.12 & 52.06 & 56.25 & 71.88 & 71.14 \\
        CL-Anomaly& 95.24 & 94.81 & 96.58 & \textbf{91.42} & 91.75 & \textbf{83.97} & \underline{71.43} & \textbf{91.71} & \textbf{89.61} \\
        \bottomrule
    \end{tabular}}
\end{table}

\begin{table}[h]
    \centering
    \caption{Recall of each method across all tasks.}
    \label{ap-tab:rec}
    \resizebox{\linewidth}{!}{
    \begin{tabular}{l cccc cccc c}
        \toprule
        Method & WebAD-1 & WebAD-2 & WebAD-3 & Mvtec-AD & BMAD & Mvtec-3D & Real3D-AD & Anomaly-Shapenet & \textit{Average}  \\
        \midrule
        Zero-shot & 73.00 & 73.24 & 72.38 & 88.05 & 70.14 & 62.96 & 66.15 & 17.19 & 65.39 \\
        Multi-task  & 95.50 & 94.84 & 96.46 & 95.18 & 96.67 & 71.43 & 63.04 & 57.55 & 83.83 \\
        \midrule
        LoRA & 63.55 & 64.08 & 64.09 & 80.92 & 59.72 & 64.55 & \underline{74.32} & \textbf{58.07} & 66.16 \\
        MoELoRA  & 74.35 & 71.83 & 74.36 & \underline{84.91} & 64.72 & \textbf{71.69} & 27.63 & 49.22 & 64.84 \\
        CL-MoE & 77.73 & 76.29 & \underline{77.57} & \textbf{86.37} & 69.58 & \underline{69.84} & 32.68 & 40.36 & 66.30 \\
        HiDe-LLaVA & \underline{78.74} & \underline{79.34} & 76.9 & 75.68 & \textbf{96.25} & 46.83 & 52.53 & 35.94 & \underline{67.75} \\
        CL-Anomaly& \textbf{87.85} & \textbf{83.57} & \textbf{87.29} & \underline{84.91} & \underline{94.17} & 52.65 & \textbf{81.71} & \underline{51.82} & \textbf{78.00} \\
        \bottomrule
    \end{tabular}}
\end{table}

\begin{table}[h]
    \centering
    \caption{F1-score of each method across all tasks.}
    \label{ap-tab:f1}
    \resizebox{\linewidth}{!}{
    \begin{tabular}{l cccc cccc c}
        \toprule
        Method & WebAD-1 & WebAD-2 & WebAD-3 & Mvtec-AD & BMAD & Mvtec-3D & Real3D-AD & Anomaly-Shapenet & \textit{Average}  \\
        \midrule
        Zero-shot & 83.53 & 83.20 & 83.02 & 85.11 & 75.54 & 57.07 & 60.39 & 28.33 & 69.52 \\
        Multi-task  & 93.30 & 92.03 & 93.92 & 95.58 & 95.67 & 74.38 & 74.31 & 70.27 & 86.18 \\
        \midrule
        LoRA & 76.82 & 76.90 & 77.23 & 84.65 & 72.15 & \textbf{66.76} & \underline{72.35} & \textbf{71.02} & \underline{74.74} \\
        MoELoRA  & 84.58 & 82.42 & 84.65 & 86.91 & 76.02 & \underline{64.91} & 40.57 & 63.53 & 72.95 \\
        CL-MoE & \underline{86.75} & \underline{85.30} & \underline{86.61} & \underline{87.66} & 79.33 & 63.46 & 43.75 & 54.87 & 73.47 \\
        HiDe-LLaVA & 75.39 & 75.49 & 75.11 & 78.22 & \textbf{93.08} & 49.30 & 54.33 & 47.92 & 68.61 \\
        CL-Anomaly& \textbf{91.40} & \textbf{88.83} & \textbf{91.70} & \textbf{88.04} & \underline{92.94} & 64.72 & \textbf{76.23} & \underline{66.22} & \textbf{82.51} \\
        \bottomrule
    \end{tabular}}
\end{table}

\section{Experimental Details and Additional Results}
\label{apd:exp}
\subsection{Complete Metrics Compared with SOTA Methods}
\phantomsection\label{apd:d1}
The complete $Accuracy$, $Precision$, $Recall$, and $F1\mathchar`\-score$ results for each method across all tasks are provided in the Table~\ref{ap-tab:acc} -~\ref{ap-tab:f1}. CL-Anomaly not only achieves superior overall performance in terms of $Accuracy$ but also consistently outperforms state-of-the-art methods in $Precision$, $Recall$, and $F1\mathchar`\-score$, attaining the best or second-best performance  across the majority of scenarios. Although our method does not always lead in $Precision$ on certain datasets, the slight trade-off in $Precision$ yields substantial gains in $Recall$. Given that $Precision$ and $Recall$ are typically considered jointly in anomaly detection tasks, the $F1\mathchar`\-score$ provides a more comprehensive assessment by balancing both metrics. From this perspective, CL-Anomaly demonstrates clear superiority.


\subsection{Details of Anomaly Reasoning}
\begin{figure}[t]
\centering
  \includegraphics[width=\linewidth ]
  {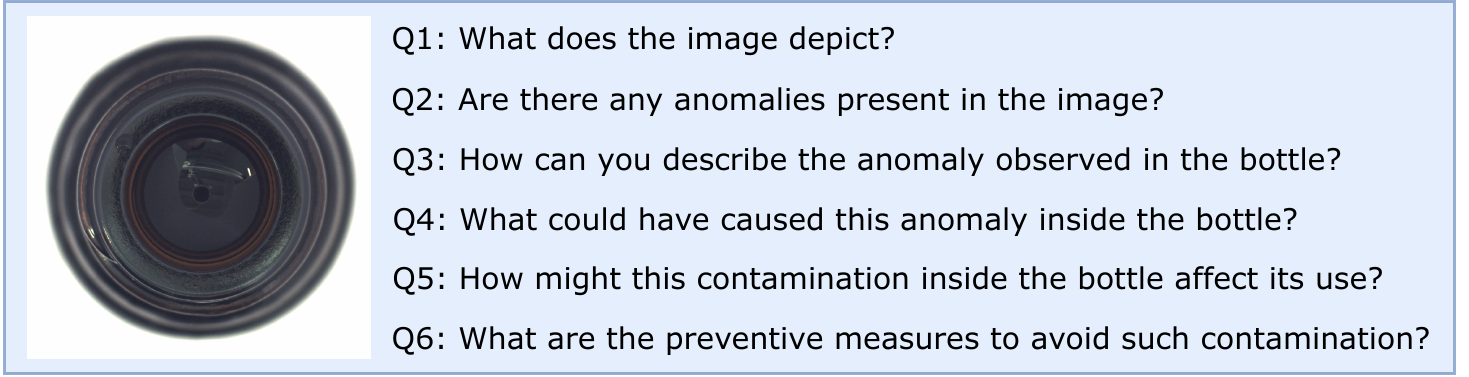}
  \caption{Anomaly reasoning question example.}
  \label{apd:case-q}
\end{figure}

\phantomsection\label{apd:d2}
\noindent\textbf{Anomaly Reasoning Question. }Anomaly Reasoning questions include image understanding, reasoning about the presence of anomalies, and interpretive reasoning of structure and attributes, etc., as exemplified in \cref{apd:case-q}.

\begin{figure}[t]
\centering
  \includegraphics[width=\linewidth ]
  {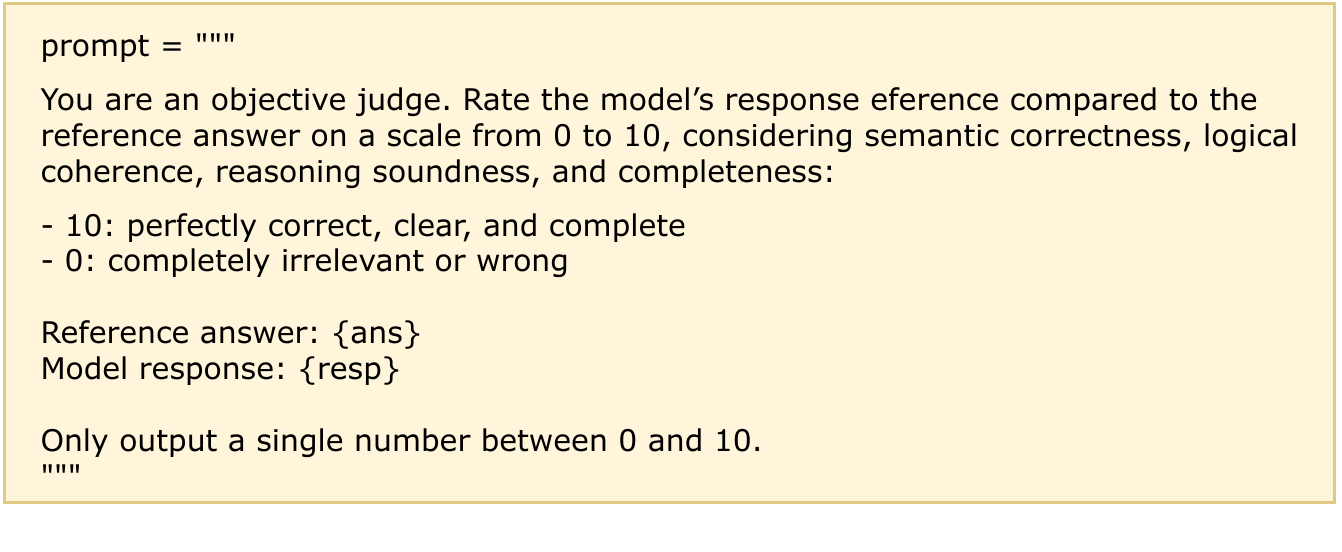}
  \caption{Prompt for LLM-Score Evaluation.}
  \label{apd:prompt}
\end{figure}

\noindent\textbf{Anomaly Reasoning Metrics. }
We adopt Rouge-L, SBERT, and LLM-Score as anomaly reasoning metrics to evaluate the model's reasoning capability from complementary perspectives. Rouge-L measures lexical overlap and sequential consistency between generated and reference texts based on longest common subsequence, assessing sentence-level structural similarity. SBERT evaluates semantic-level fidelity by computing the distance between generated and reference reasoning in the semantic embedding space. LLM-Score leverages a powerful large language model as the judge to assess reasoning quality against predefined criteria, enabling a more holistic evaluation of reasoning results. In this study, we employ Qwen-3-Max as the judging model for LLM-Score, with the detailed scoring prompt provided in \cref{apd:prompt}.

\begin{table}[ht]
    \centering
    \caption{Detailed metric comparison with SOTA continual learning approaches in traditional continuous anomaly detection scenarios.}
    \resizebox{\linewidth}{!}{
    \begin{tabular}
        {cl ccccc ccccc ccccc |c}
        \toprule
           & {\hspace{1em}} Method & zipper & hazelnut & leather & cable & metal nut & wood & transistor & tile & screw & carpet & bottle & capsule & toothbrush & pill & grid  & \textit{Average} \\
  
        \midrule
            \multirow{4}{*}{\rotatebox{90}{\textit{Accuracy}}} 
            & {\hspace{1em}}MoE-LoRA & 97.81 & 95.24 & 100 & 92.25 & 96.64 & 98.20 & 88.79 & 98.32 & 88.17 & 99.28 & 70.30 & 91.74 & 97.30 & 85.81 & 100 & \underline{93.32}  \\ 
            & {\hspace{1em}}HiDe-LLaVA & 90.51 & 72.02 & 92.86 & 74.42 & 88.24 & 92.79 & 58.88 & 98.32 & 82.25 & 97.83 & 61.39 & 90.91 & 91.89 & 92.90 & 100  & 85.68 \\ 
            & {\hspace{1em}}CL-MoE & 97.08 & 91.67 & 100 & 91.47 & 94.96 & 98.20 & 86.92 & 98.32 & 86.39 & 99.28 & 59.41 & 90.08 & 94.59 & 79.35 & 100 & 91.18 \\  
            & {\hspace{1em}}CL-Anomaly & 97.81 & 95.83 & 100 & 91.47 & 98.32 & 100 & 86.92 & 98.32 & 86.39 & 100 & 87.13 & 95.04 & 97.30 & 94.84 & 100 & \textbf{95.29} \\  
            \midrule

            \multirow{4}{*}{\rotatebox{90}{\textit{Precision}}} 
            & {\hspace{1em}}MoE-LoRA & 95.56 & 78.38 & 100 & 88.57 & 96.15 & 91.67 & 52.38 & 100 & 74.58 & 97.50 & 32.56 & 89.47 & 100 & 68.85 & 100 & \underline{84.38} \\ 
            & {\hspace{1em}}HiDe-LLaVA & 96.97 & 38.16 & 81.25 & 58.33 & 71.88 & 75.00 & 18.37 & 95.00 & 80.65 & 92.86 & 27.78 & 100 & 100 & 100 & 100 & 75.75 \\ 
            & {\hspace{1em}}CL-MoE & 95.45 & 67.44 & 100 & 88.57 & 95.83 & 91.67 & 47.83 & 100 & 70.97 & 97.50 & 25.93 & 86.84 & 87.50 & 59.42 & 100 & 80.99\\  
            & {\hspace{1em}}CL-Anomaly & 95.56 & 80.56 & 100 & 86.11 & 100 & 100 & 47.83 & 100 & 71.67 & 100 & 54.17 & 94.74 & 100 & 91.11 & 100 & \textbf{88.12} \\  
            \midrule

            \multirow{4}{*}{\rotatebox{90}{\textit{Recall}}} 
            & {\hspace{1em}}MoE-LoRA & 97.73 & 100 & 100 & 83.78 & 89.47 & 100 & 84.62 & 94.74 & 89.80 & 100 & 93.33 & 85.00 &87.50 & 93.33 & 100 & \textbf{93.29} \\ 
            & {\hspace{1em}}HiDe-LLaVA & 72.73 & 100 & 100 & 37.84 & 82.14 & 95.45 & 69.23 & 100 & 51.02 & 100 & 100 & 72.50 & 62.50 & 75.56 & 100 & 81.26 \\ 
            & {\hspace{1em}}CL-MoE & 95.45 & 100 & 100 & 83.78 & 82.14 & 100 & 84.62 & 94.74 & 89.80 & 100 & 93.33 & 82.50 & 87.50 & 91.11 & 100 & 92.33 \\  
            & {\hspace{1em}}CL-Anomaly & 97.73 & 100 & 100 & 83.78 & 92.86 & 100 & 84.62 & 94.74 & 87.76 & 100 & 86.67 & 90.00 & 87.50 & 91.11 & 100 & \underline{93.12}\\  
            \midrule
            

            \multirow{4}{*}{\rotatebox{90}{\textit{F1-score}}} 
            & {\hspace{1em}}MoE-LoRA & 96.63 & 87.88 & 100 & 86.11 & 92.59 & 95.65 & 64.71 & 97.30 & 81.48 & 98.73 & 48.28 & 87.18 & 93.33 & 79.25 & 100 & \underline{87.27}  \\ 
            & {\hspace{1em}}HiDe-LLaVA & 83.12 & 55.24 & 89.66 & 45.90 & 76.67 & 84.00 & 29.03 & 97.44 & 62.50 & 96.30 & 43.48 & 84.06 & 76.92 & 86.08 & 100  & 74.03 \\ 
            & {\hspace{1em}}CL-MoE & 95.45 & 80.56 & 100 & 86.11 & 88.46 & 95.65 & 61.11 & 97.30 & 79.28 & 98.73 & 40.58 & 84.62 & 87.50 & 71.93 & 100 & 84.49 \\  
            & {\hspace{1em}}CL-Anomaly & 96.63 & 89.23 & 100 & 84.93 & 96.30  & 100 & 61.11 & 97.30 & 78.90 & 100 & 66.67 & 92.31 & 93.33 & 91.11 & 100 & \textbf{89.85} \\  
        \bottomrule
    \end{tabular}}
    \label{tab:pra_ad}
\end{table}

\subsection{Details of Traditional Continuous AD Scenarios}
\phantomsection\label{apd:d3}

\noindent\textbf{Experimental Setup. }
In previous experiments, we demonstrated the continual learning capability of CL-Anomaly in complex multi-domain and multi-modal anomaly detection scenarios. To align with the experimental settings of traditional continual anomaly detection, we follow the experimental setup of \cite{tang2024incremental,liu2024unsupervised} to validate the model's continual anomaly detection ability in single-anomaly scenarios. Under this setting, the task of each stage is to learn a single category (e.g., Bottle or Capsule in MVTec-AD), rather than an entire dataset (e.g., WebAD or BMAD), forming a basic class-incremental learning scenario.

\noindent\textbf{Detailed Experimental Results. }
We conduct experiments on MVTec-AD, a classic anomaly detection dataset, adopting the random sequence Zipper $\rightarrow$ Hazelnut $\rightarrow$ Leather $\rightarrow$ $\dots$ $\rightarrow$ Pill $\rightarrow$ Grid, as shown in Table~\ref{tab:pra_ad}. The model is trained in a sequential continual learning manner, where each category is learned incrementally, and the performance for each category is evaluated after completing training on the entire sequence. The complete results are reported in Table~\ref{tab:pra_ad}. Under this single-scenario continual anomaly detection setting, our method achieves superior overall performance across all metrics compared to state-of-the-art continual learning approaches. Although Recall is not consistently the highest, it should be jointly assessed with Precision; notably, CL-Anomaly demonstrates clear advantages in F1-score, which comprehensively balances both metrics.

\subsection{Further Experiments in CIL}
\begin{figure}[ht]
    \centering
    \begin{subfigure}[b]{0.48\linewidth}
        \centering
        \includegraphics[width=\linewidth]{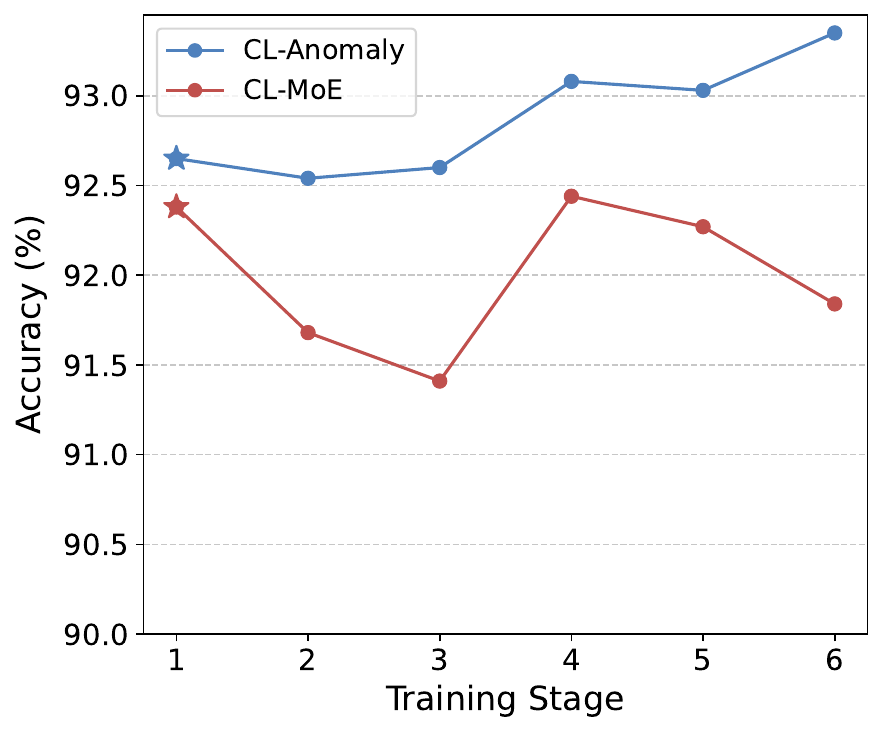}
        \caption{Comparison of CIL results on WebAD-1}
        \label{apd:web1}
    \end{subfigure}
    \hfill
    \begin{subfigure}[b]{0.48\linewidth}
        \centering
        \includegraphics[width=\linewidth]{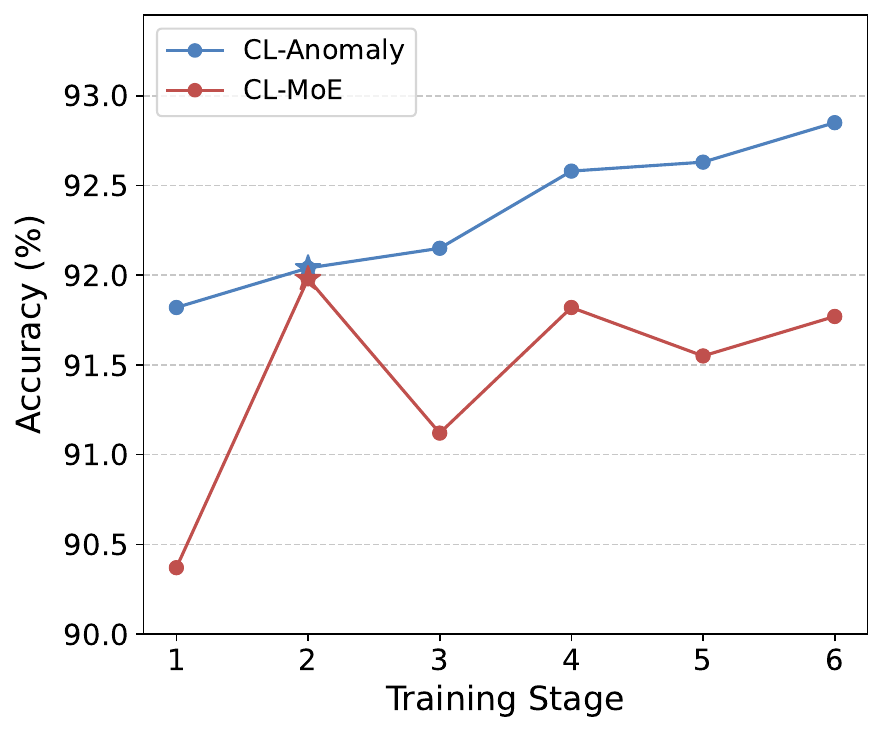}
        \caption{Comparison of CIL results on WebAD-2}
        \label{apd:web2}
    \end{subfigure}
    
    \vspace{2mm} 
    
    \begin{subfigure}[b]{0.48\linewidth}
        \centering
        \includegraphics[width=\linewidth]{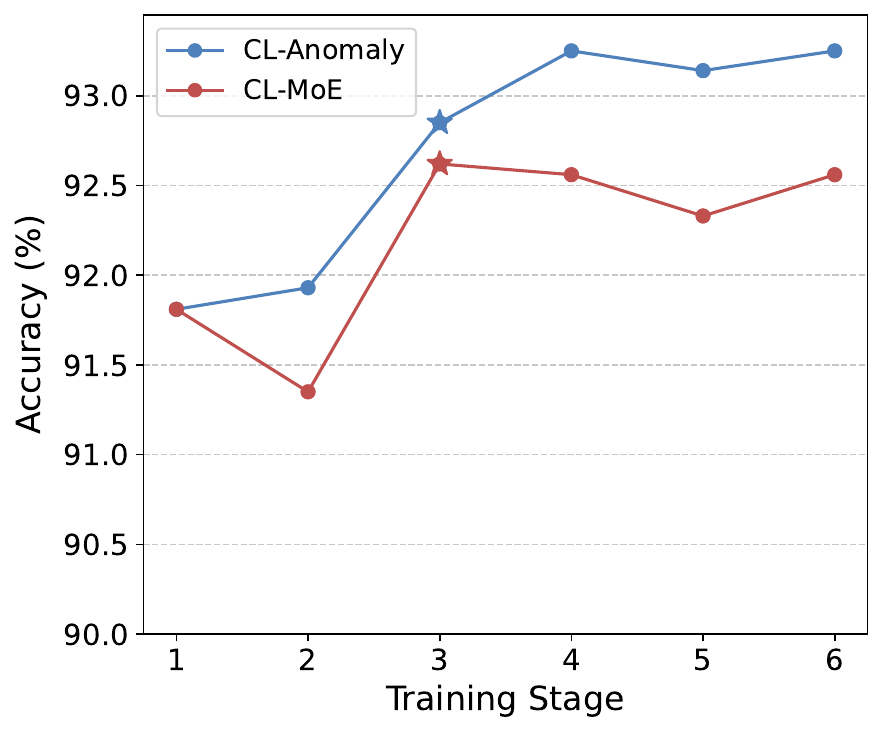}
        \caption{Comparison of CIL results on WebAD-3}
        \label{apd:web3}
    \end{subfigure}
    \hfill
    \begin{subfigure}[b]{0.48\linewidth}
        \centering
        \includegraphics[width=\linewidth]{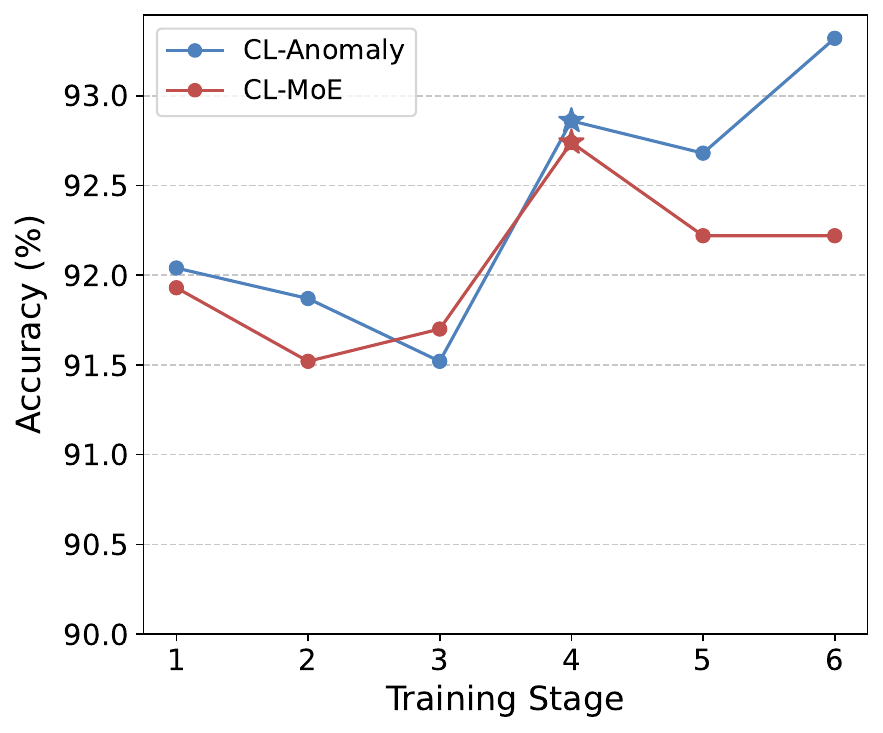}
        \caption{Comparison of CIL results on WebAD-4}
        \label{apd:web4}
    \end{subfigure}

    \vspace{2mm} 

    \begin{subfigure}[b]{0.48\linewidth}
        \centering
        \includegraphics[width=\linewidth]{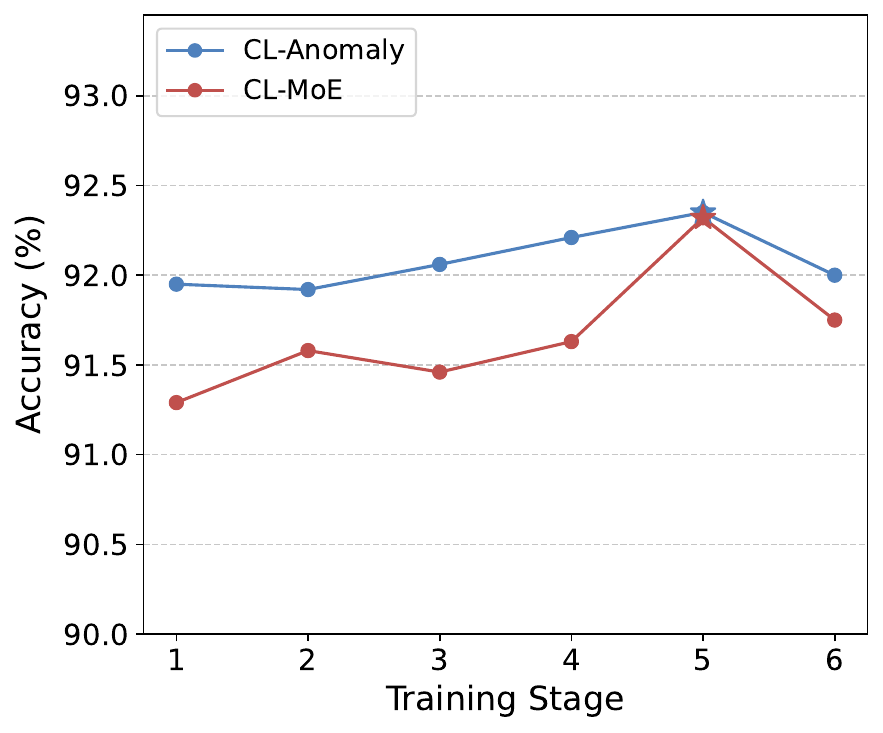}
        \caption{Comparison of CIL results on WebAD-5}
        \label{apd:web5}
    \end{subfigure}
    \hfill
    \begin{subfigure}[b]{0.48\linewidth}
        \centering
        \includegraphics[width=\linewidth]{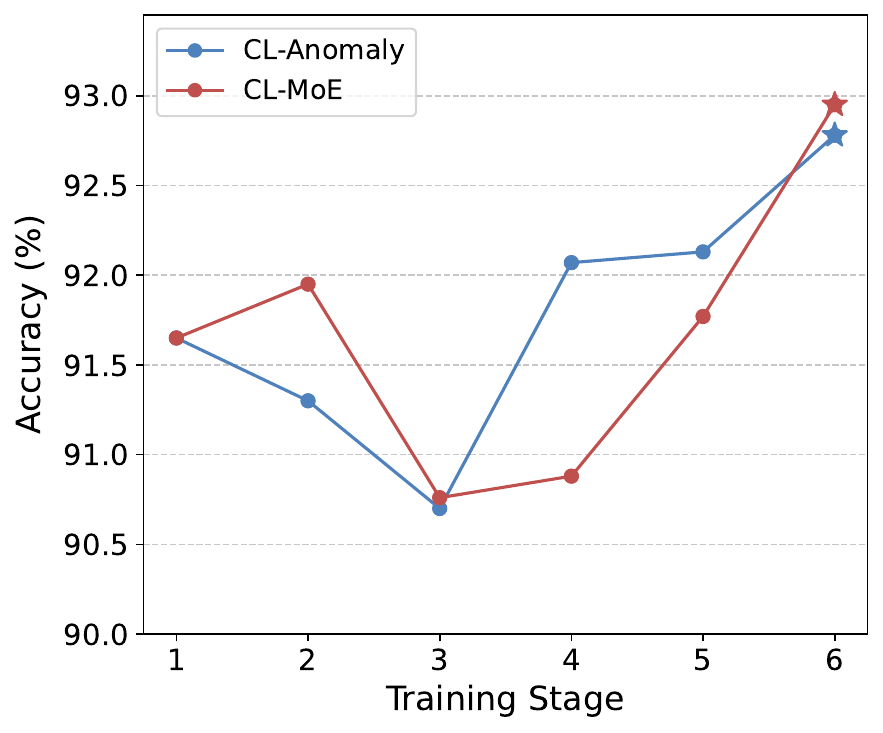}
        \caption{Comparison of CIL results on WebAD-6}
        \label{apd:web6}
    \end{subfigure}

    \caption{Comparison with CL-MoE in CIL across WebAD datasets. Each subfigure corresponds to one dataset.}
    \label{fig:webad_cil_all}
\end{figure}

We further provide additional comparison results between CL-Anomaly and CL-MoE in class-incremental learning (CIL), using the same experimental setting as the \textit{Analysis of Knowledge Transfer in CIL in section 5.4} of the main paper. As shown in \cref{apd:web1}-\cref{apd:web6}, our method demonstrates consistently strong overall performance across all stages. Notably, the performance gap is most salient for tasks learned earlier in the sequence, highlighting CL-Anomaly’s strong resistance to catastrophic forgetting and its positive knowledge transfer in class-incremental learning for anomaly detection.

\clearpage

\clearpage

\begin{figure}[ht]
    \centering
    \begin{subfigure}[b]{0.48\linewidth}
        \centering
        \includegraphics[width=\linewidth]{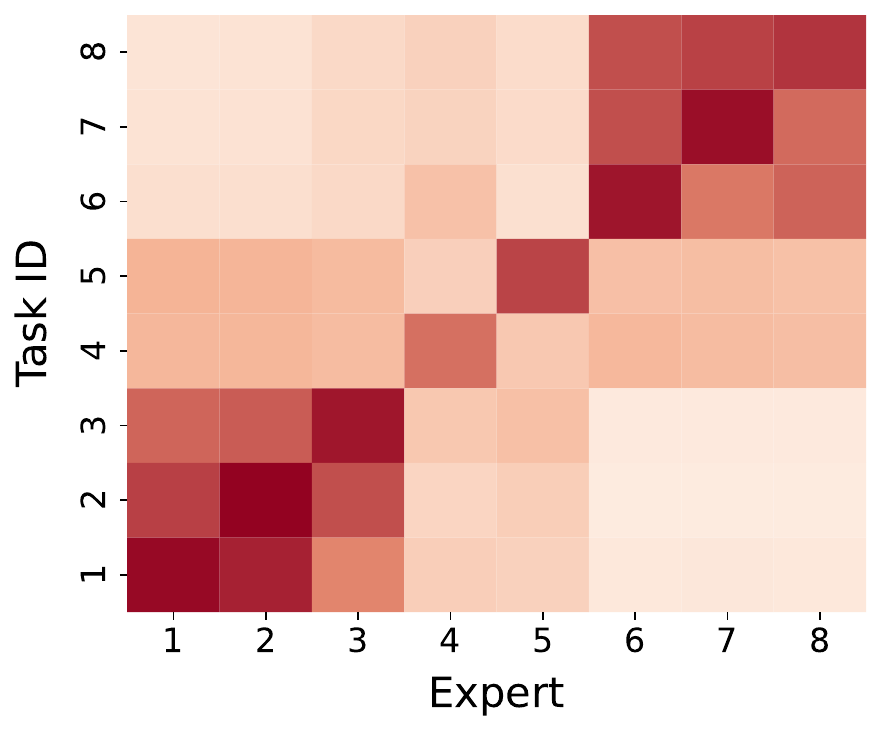}
        \caption{Visual-anchor routing selection}
        \label{fig:priv_router_heat} 
    \end{subfigure}
    \hfill
    \begin{subfigure}[b]{0.48\linewidth}
        \centering
        \includegraphics[width=\linewidth]{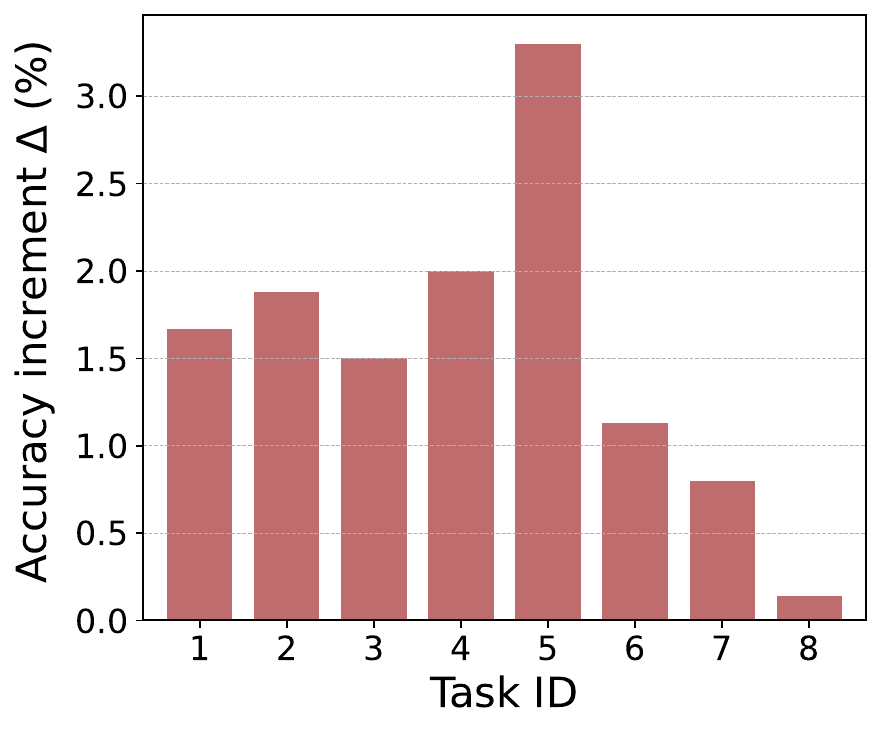}
        \caption{Fully correct routing selection}
        \label{fig:priv_router_acc} 
    \end{subfigure}
    \caption{Analysis of PrivLoRA Routing Selection. (a) Heatmap of PrivLoRA visual-anchor routing, with darker colors indicating higher expert weighting for the task. (b) Last accuracy improvement with manually corrected PrivLoRA routing.}
    \label{fig:prvilora_router}
\end{figure}

\section{Limitation}
\phantomsection\label{apd:lim}
During inference, each PrivLoRA is assigned a weight based on the cosine similarity between an input’s visual anchor and stored task anchors, such that visual-anchor routing directly impacts model performance. To assess this effect, we present the average routing heatmap across tasks after training, as shown in Figure~\ref{fig:priv_router_heat}. The results show that the heatmap is diagonally dominant, with each task’s PrivLoRA predominantly activated. However, we also observe that the first three tasks from the same dataset and the last three 3D-modality tasks form high-response clusters, suggesting that tasks with similar distributions or modalities exhibit highly consistent visual features. This suggests that in scenarios where visual features are similar but the underlying anomaly semantics differ, the routing selections of PrivLoRA may not always be sufficiently precise.

To evaluate the performance upper bound, we manually construct an ideal routing scenario by assigning $100\%$ weight to the current task’s PrivLoRA and zero to all others, as shown in Figure~\ref{fig:priv_router_acc}. Under this setting, all tasks exhibit performance gains, highlighting the considerable potential of private expert routing. Consequently, the current private expert routing mechanism is not yet fully refined. How to more efficiently leverage task-specific expert knowledge remains an open problem warranting further investigation.

\section{Visualization Examples}
\begin{figure*}[t]
    \centering

    \begin{subfigure}[b]{0.48\linewidth}
        \centering
        \includegraphics[width=\linewidth]{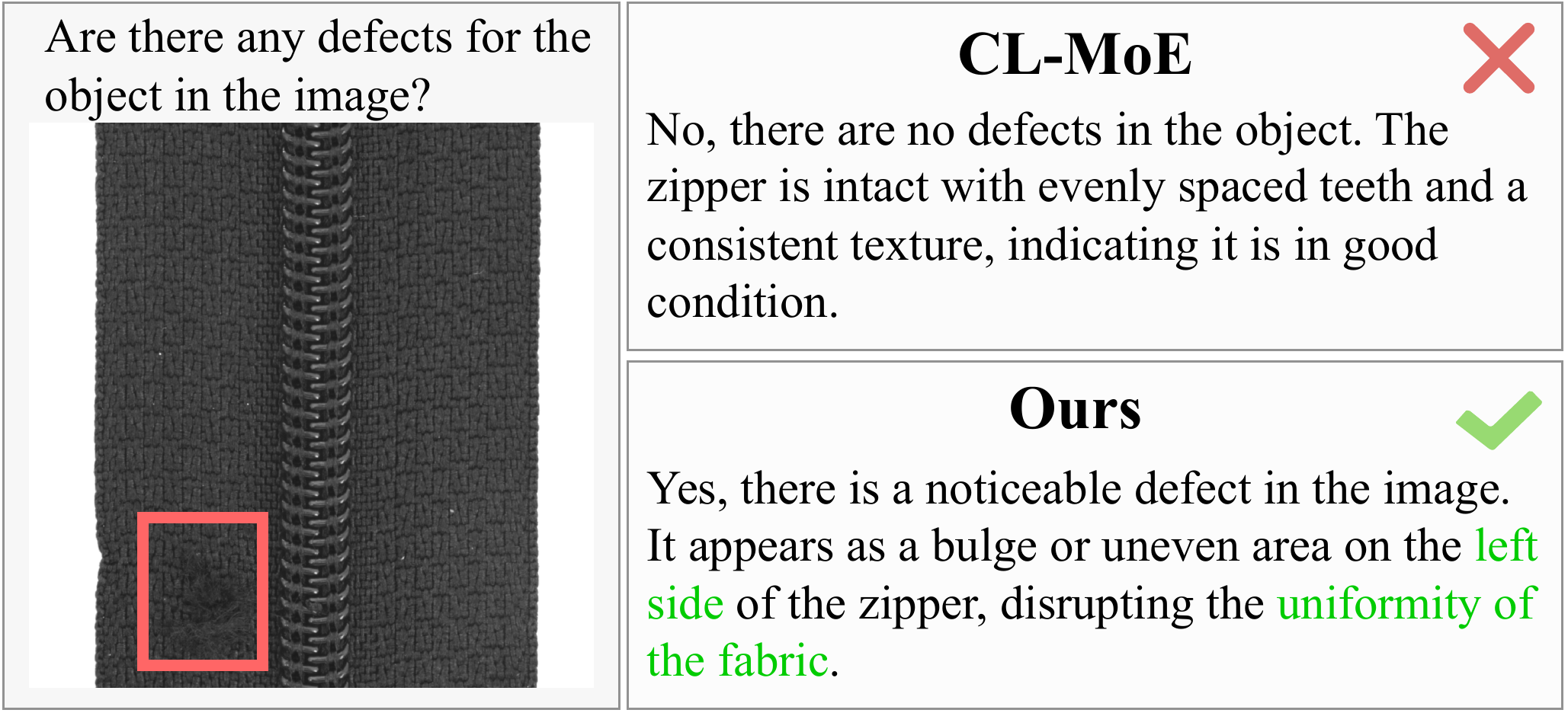}
        \caption{Mvtec-AD (Zipper)}
        \label{apd:case1}
    \end{subfigure}
    \hfill
    \begin{subfigure}[b]{0.48\linewidth}
        \centering
        \includegraphics[width=\linewidth]{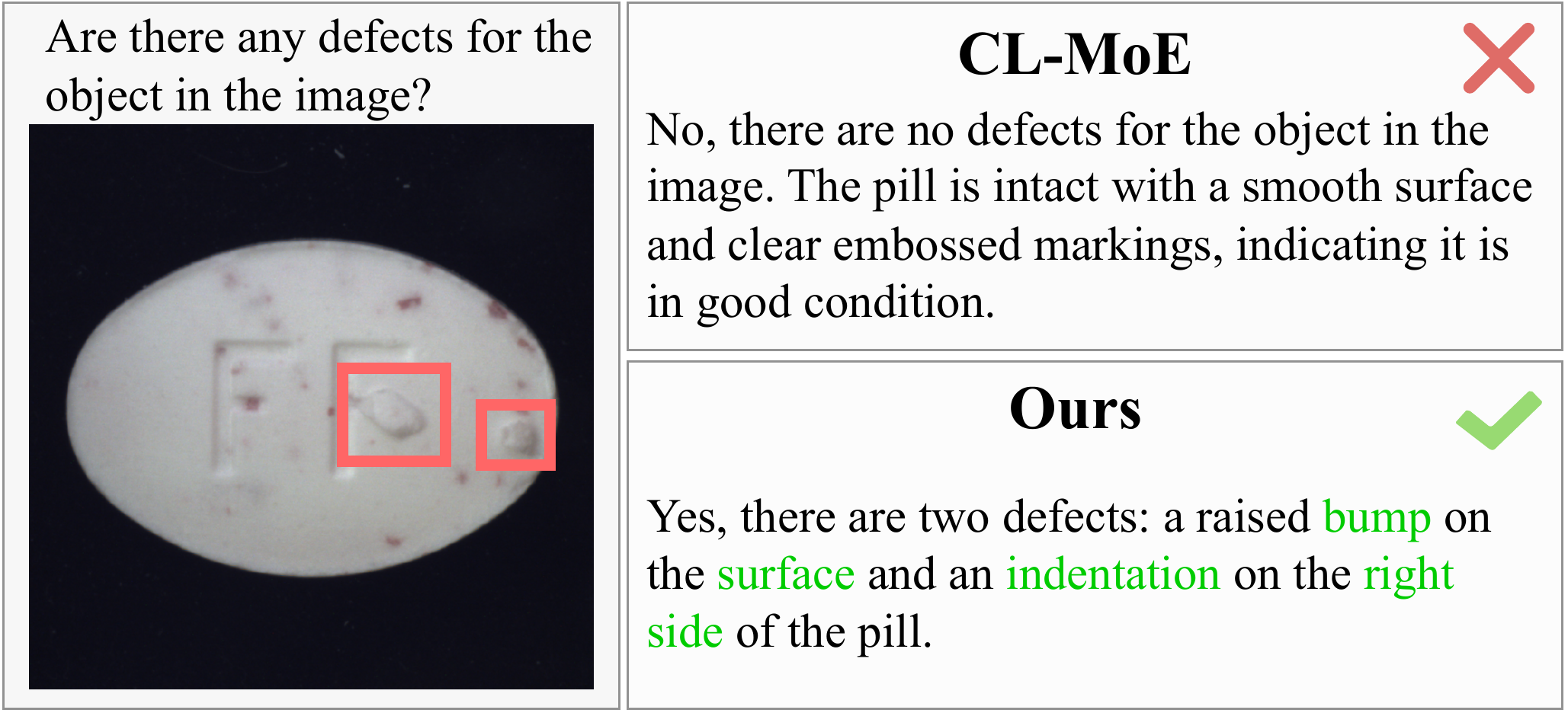}
        \caption{Mvtec-AD (Pill)}
        \label{apd:case2}
    \end{subfigure}

    \vspace{2mm} 

    \begin{subfigure}[b]{0.48\linewidth}
        \centering
        \includegraphics[width=\linewidth]{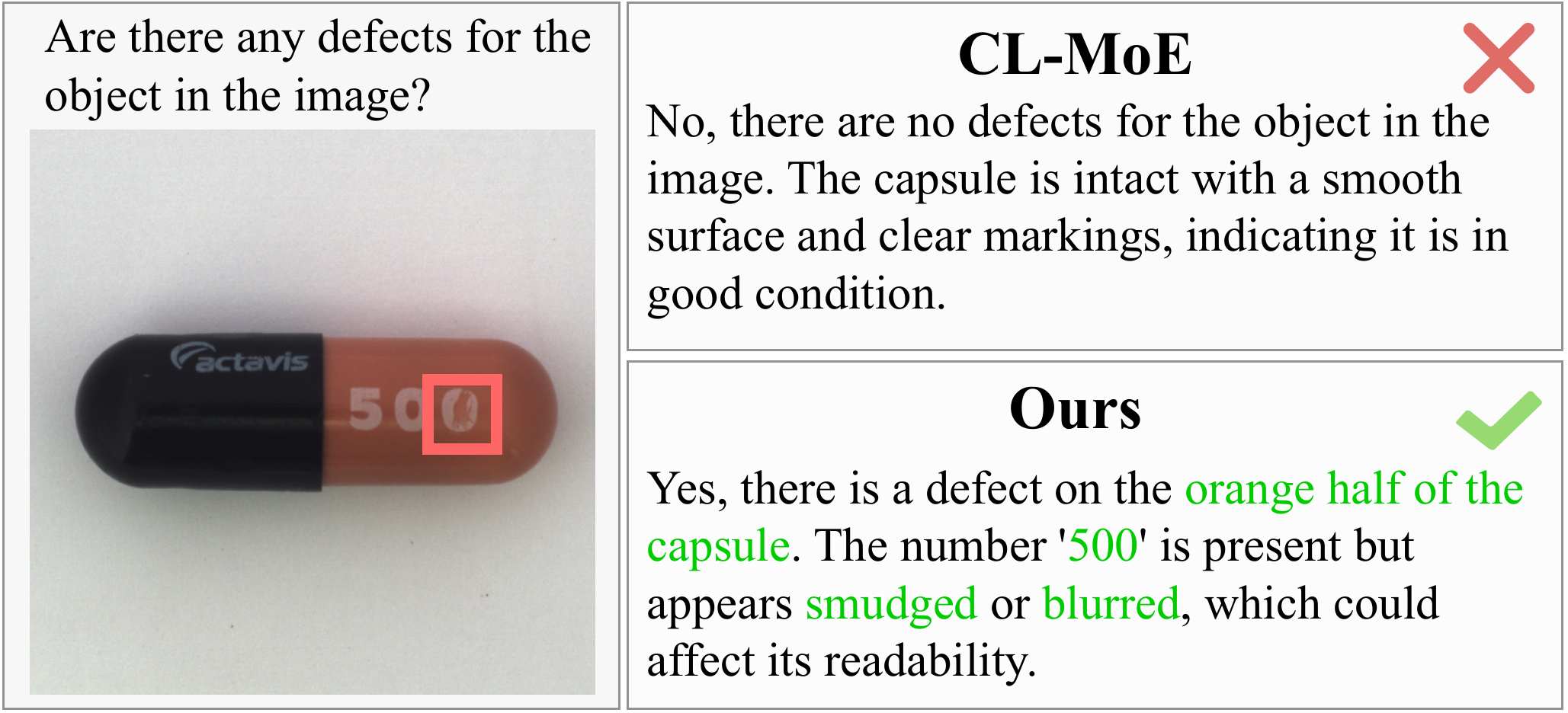}
        \caption{Mvtec-AD (Capsule)}
        \label{apd:case3}
    \end{subfigure}
    \hfill
    \begin{subfigure}[b]{0.48\linewidth}
        \centering
        \includegraphics[width=\linewidth]{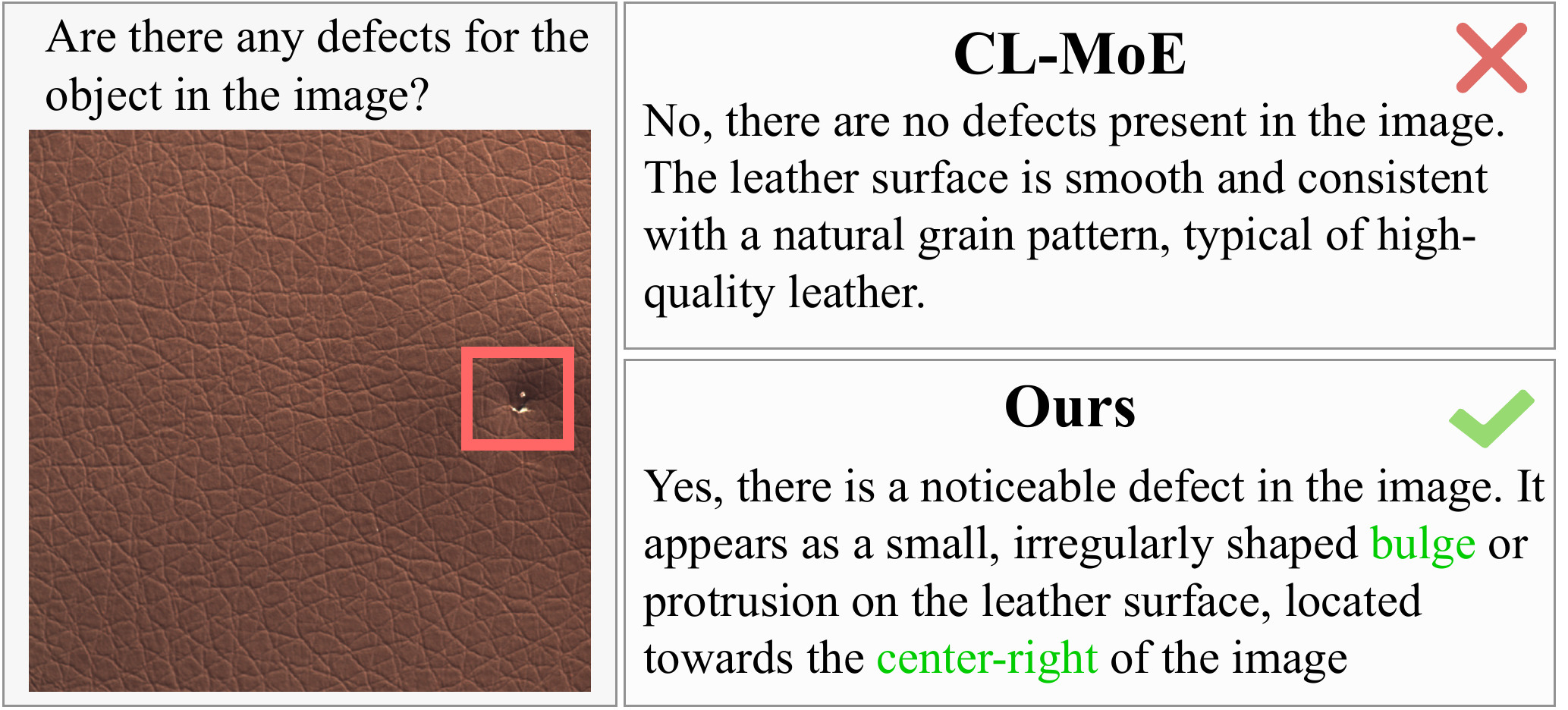}
        \caption{Mvtec-AD (Leather)}
        \label{apd:case4}
    \end{subfigure}

    \vspace{2mm} 

    \begin{subfigure}[b]{0.48\linewidth}
        \centering
        \includegraphics[width=\linewidth]{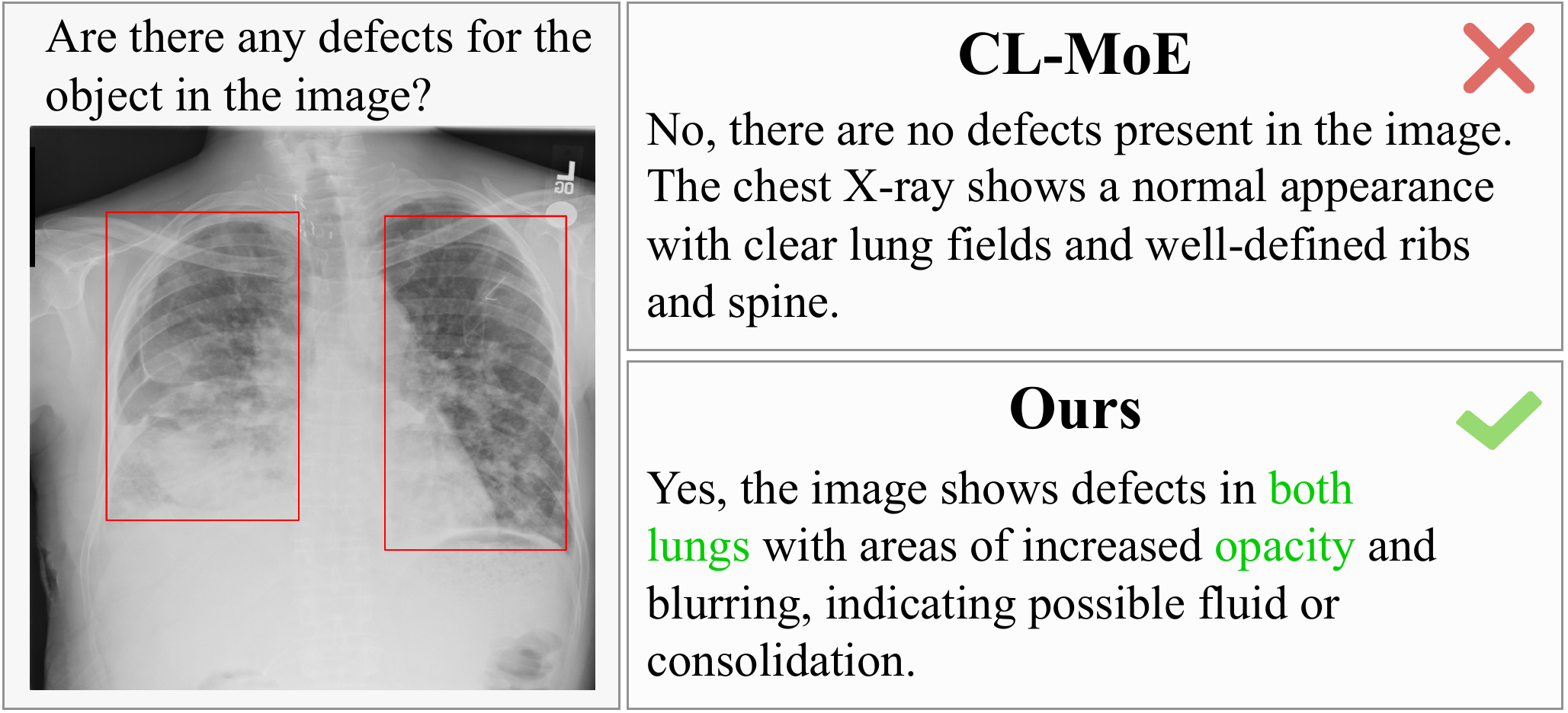}
        \caption{BMAD (Chest)}
        \label{apd:case5}
    \end{subfigure}
    \hfill
    \begin{subfigure}[b]{0.48\linewidth}
        \centering
        \includegraphics[width=\linewidth]{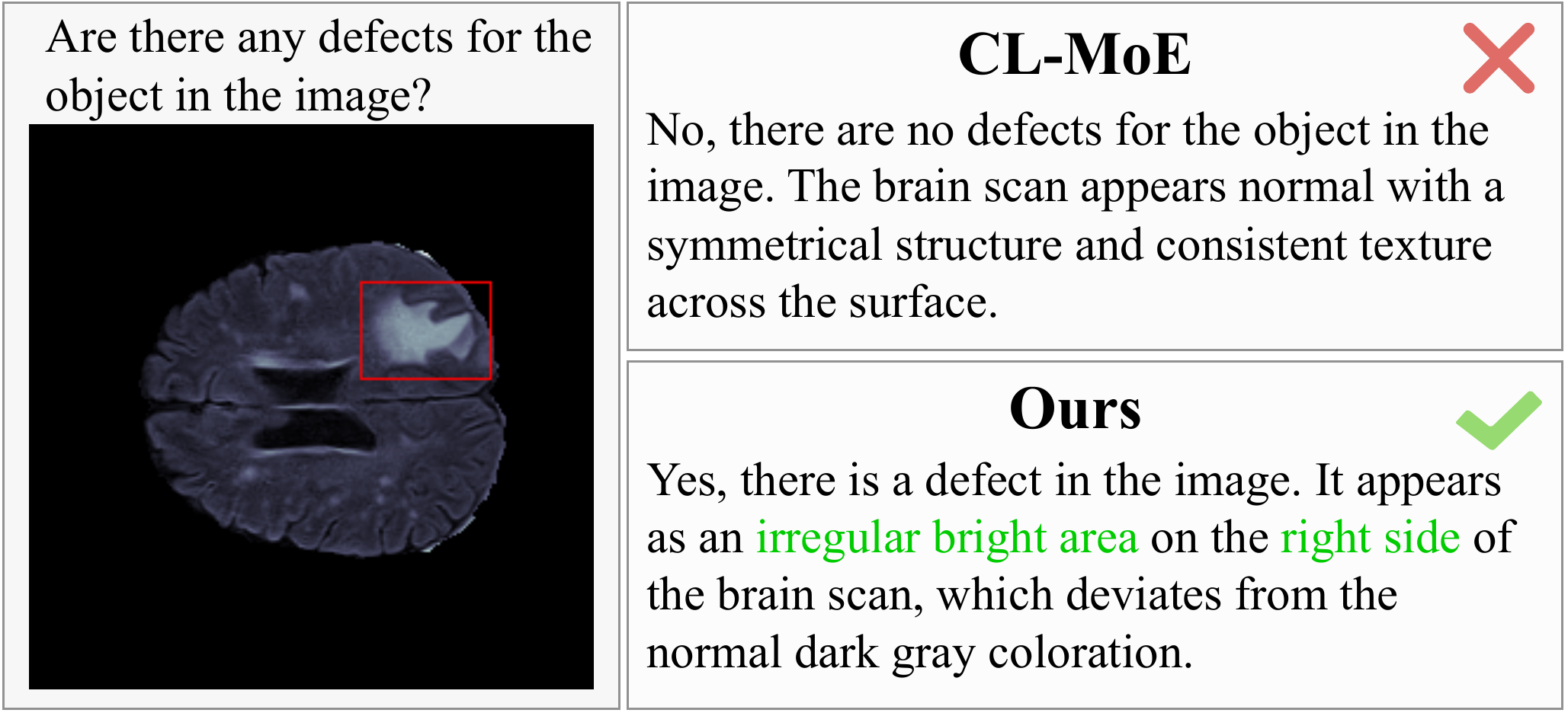}
        \caption{BMAD (Brain)}
        \label{apd:case6}
    \end{subfigure}

    \vspace{2mm} 

    \begin{subfigure}[b]{0.48\linewidth}
        \centering
        \includegraphics[width=\linewidth]{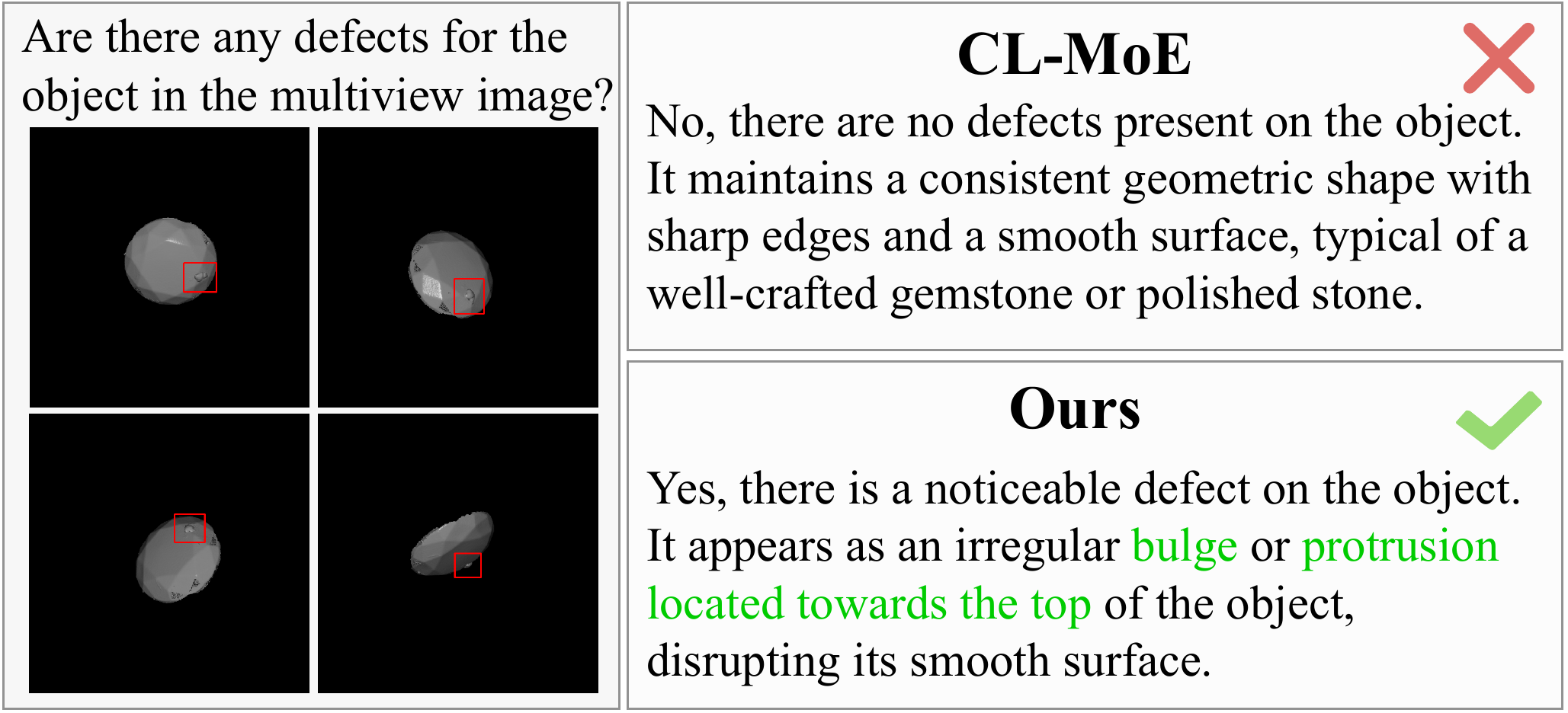}
        \caption{Real3D-AD (Diamond)}
        \label{apd:case7}
    \end{subfigure}
    \hfill
    \begin{subfigure}[b]{0.48\linewidth}
        \centering
        \includegraphics[width=\linewidth]{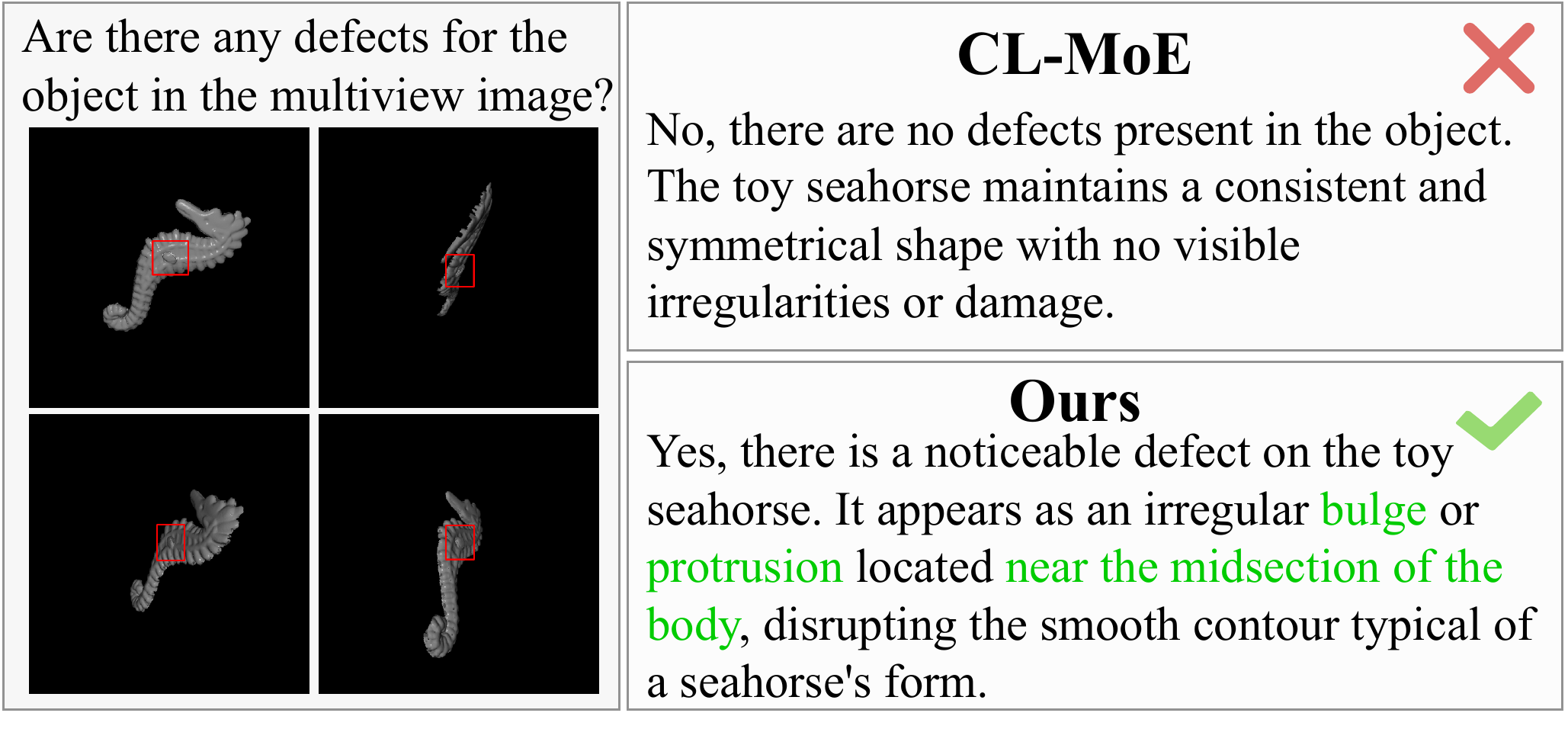}
        \caption{Real3D-AD (Seahorse)}
        \label{apd:case8}
    \end{subfigure}

    \caption{Visualization examples across datasets (Mvtec-AD, BMAD, Real3D-AD). Each subfigure corresponds to one object category.}
    \label{fig:all_cases}
\end{figure*}

\phantomsection\label{apd:vis}
\noindent\textbf{Visualization Examples of Anomaly Detection. }
We provide additional visualization example analyses (\cref{apd:case1}-\cref{apd:case8}) for anomaly detection. CL-Anomaly accurately identifies anomalous regions under simple prompting and produces clear reasoning cues (in green text), demonstrating its strong continual anomaly detection capability.

\begin{figure*}[t]
    \centering

    \begin{subfigure}[b]{0.48\linewidth}
        \centering
        \includegraphics[width=\linewidth]{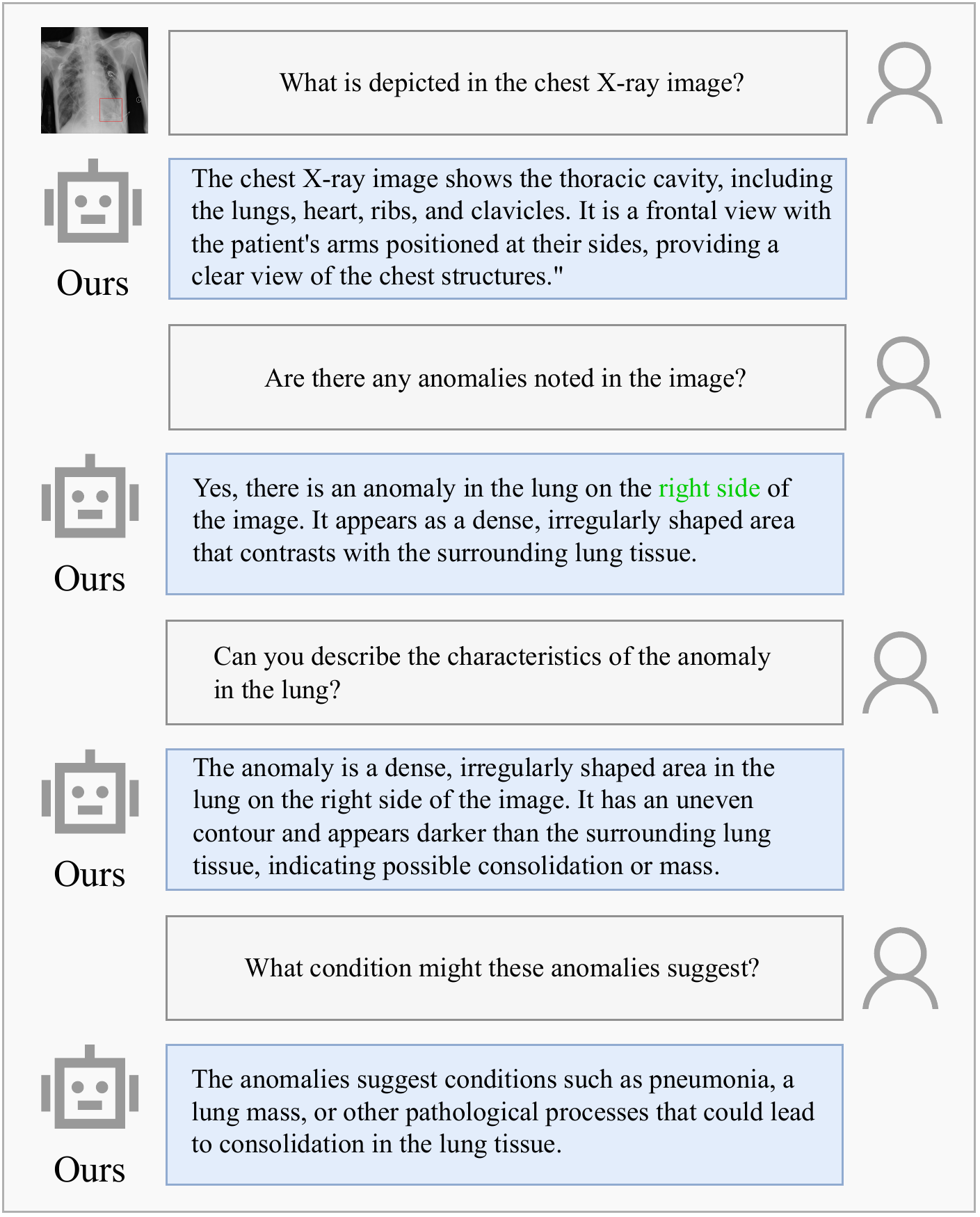}
        \caption{BMAD (Chest)}
        \label{apd:caserea-1}
    \end{subfigure}
    \hfill
    \begin{subfigure}[b]{0.48\linewidth}
        \centering
        \includegraphics[width=\linewidth]{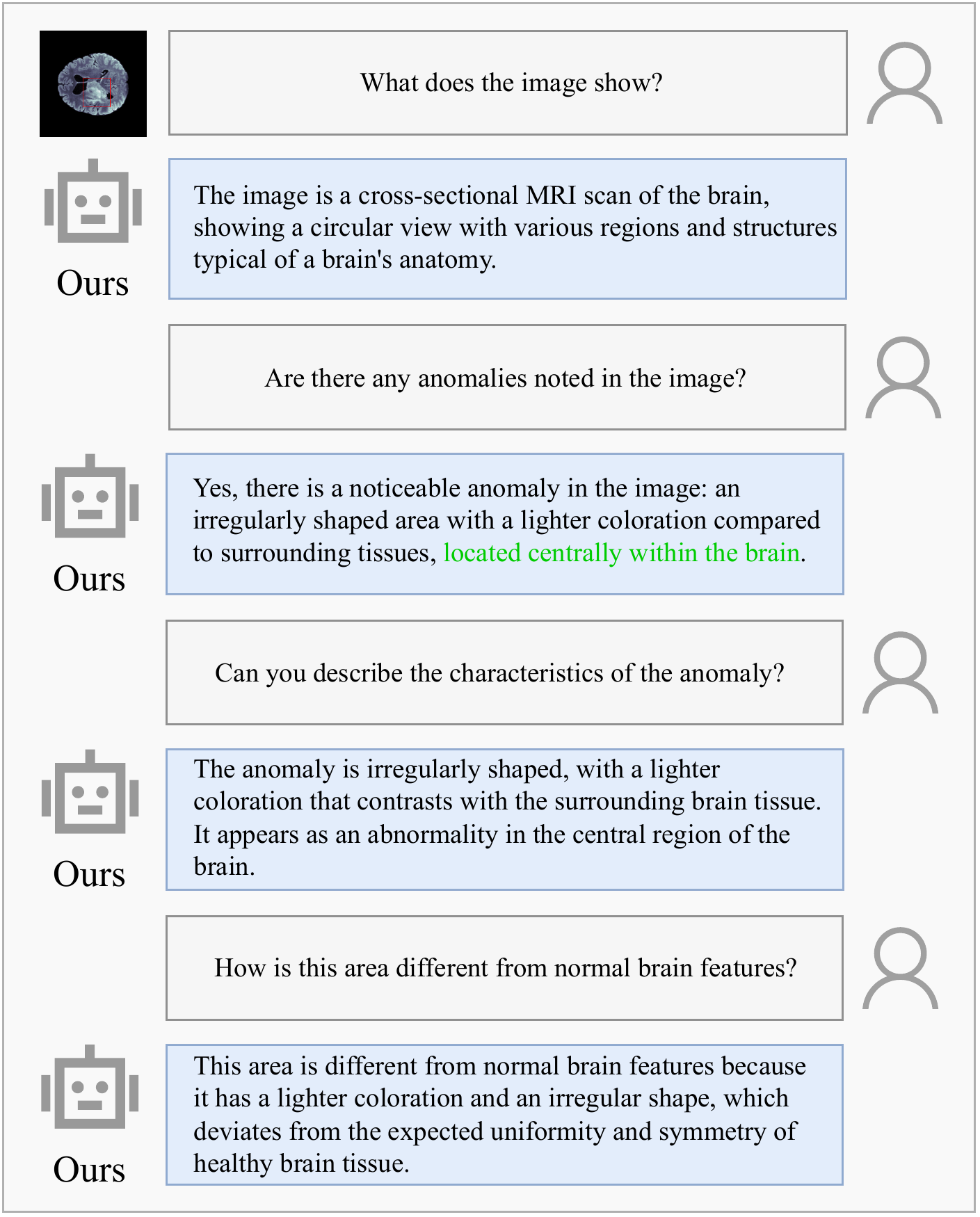}
        \caption{BMAD (Brain)}
        \label{apd:caserea-2}
    \end{subfigure}

    \vspace{2mm}

    \begin{subfigure}[b]{0.48\linewidth}
        \centering
        \includegraphics[width=\linewidth]{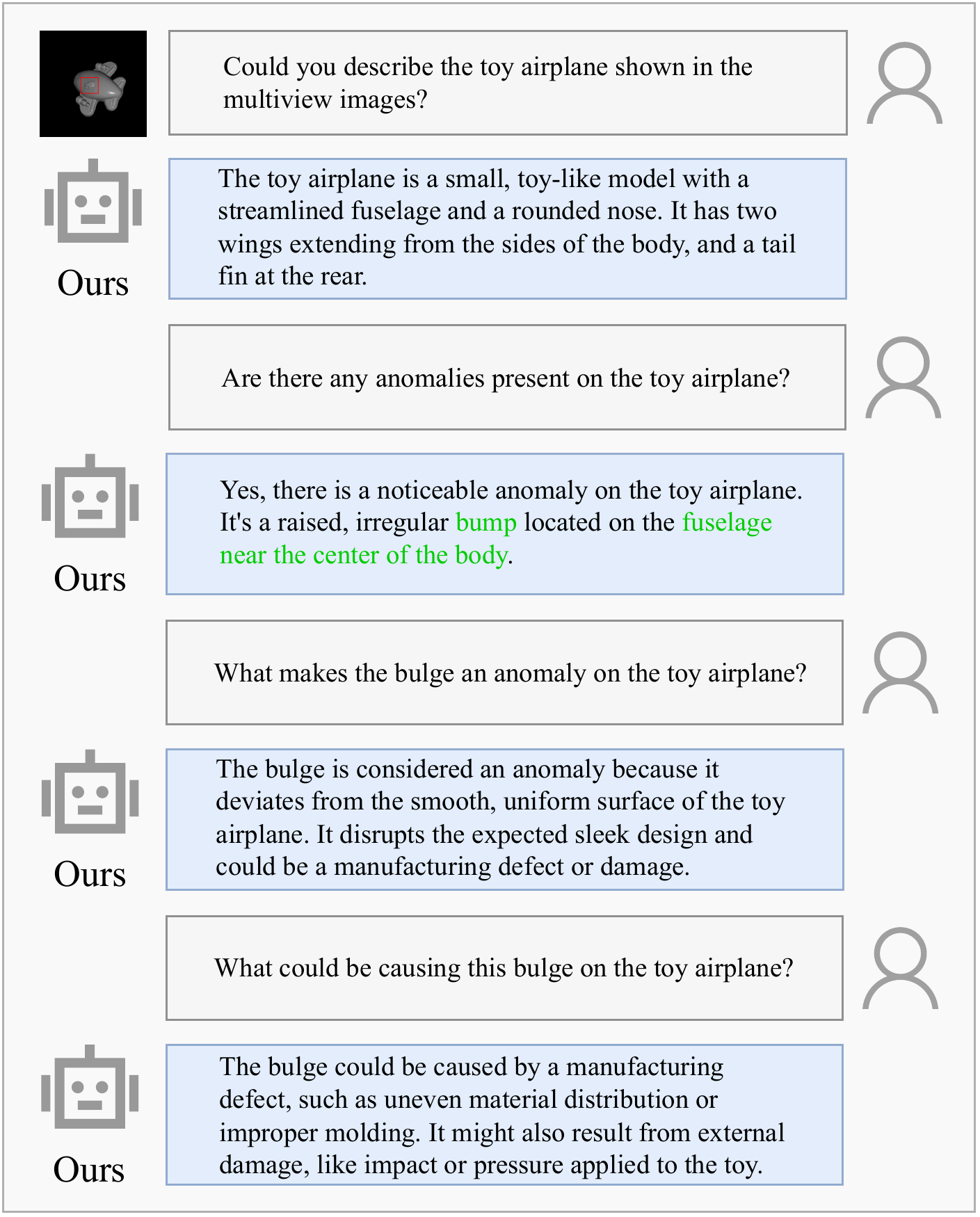}
        \caption{Real3D-AD (Airplane)}
        \label{apd:caserea-3}
    \end{subfigure}
    \hfill
    \begin{subfigure}[b]{0.48\linewidth}
        \centering
        \includegraphics[width=\linewidth]{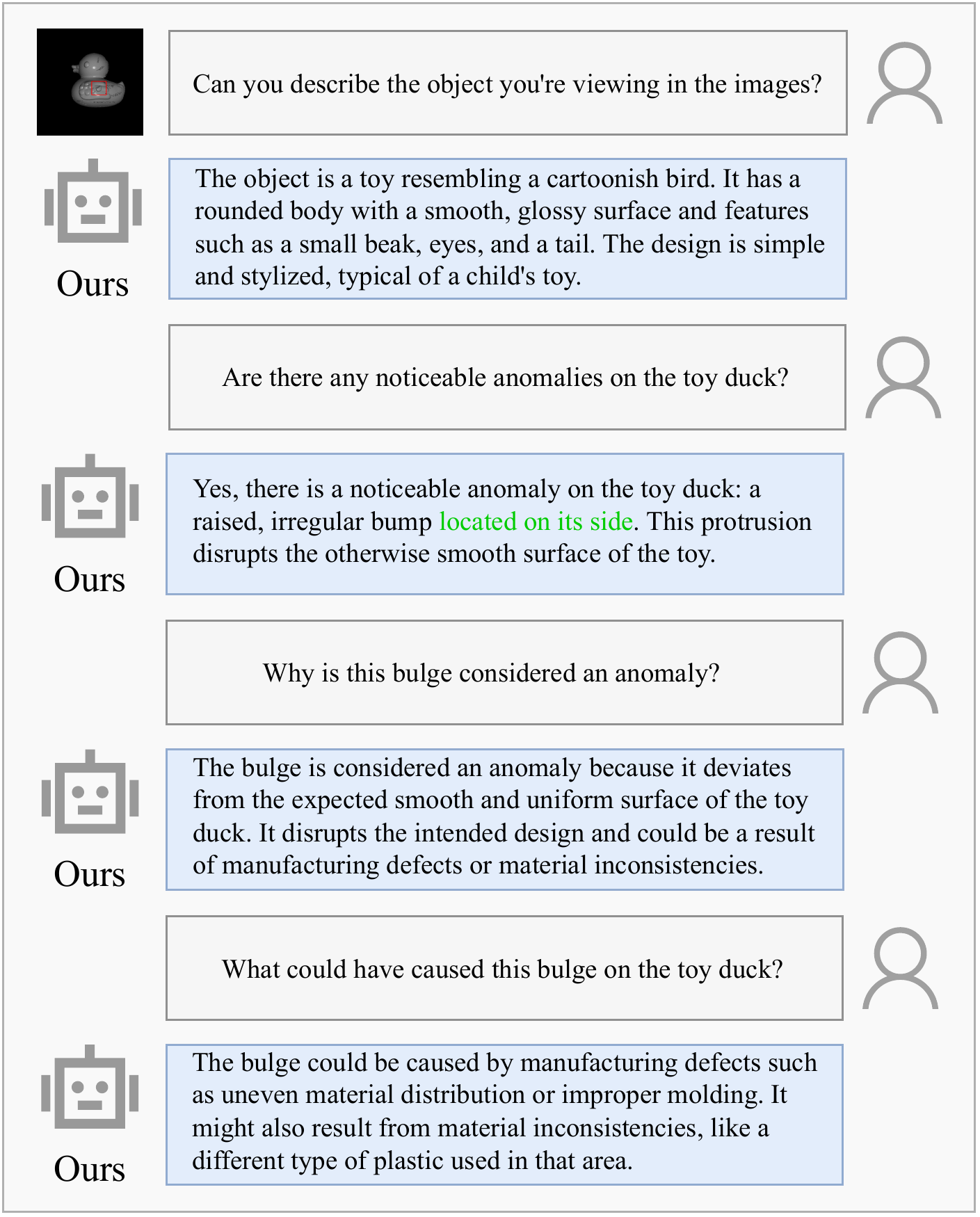}
        \caption{Real3D-AD (Duck)}
        \label{apd:caserea-4}
    \end{subfigure}

    \caption{Anomaly reasoning examples across datasets (part 1)}
    \label{fig:caserea_page1}
\end{figure*}

\begin{figure*}[t]
    \centering

    \begin{subfigure}[b]{0.48\linewidth}
        \centering
        \includegraphics[width=\linewidth]{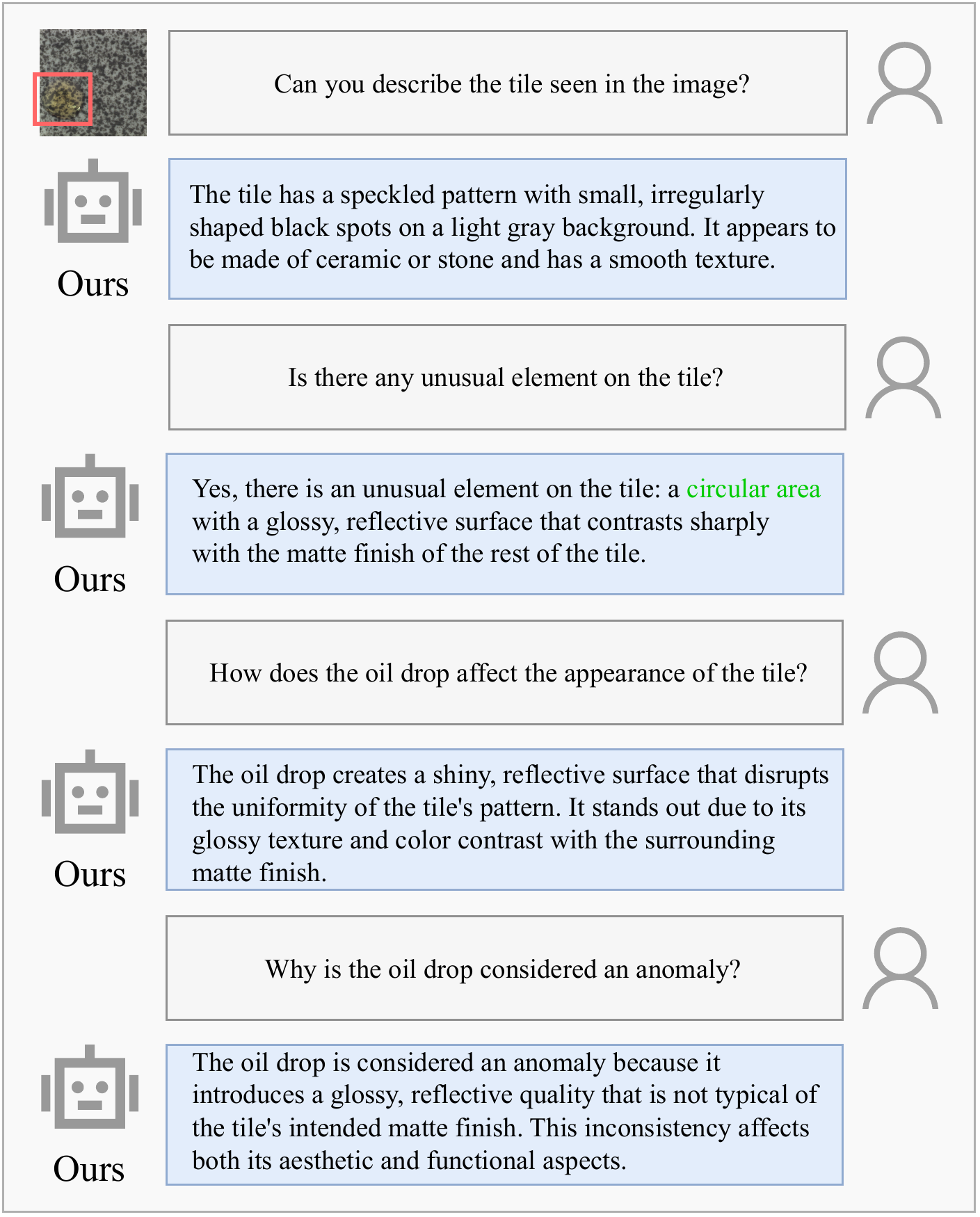}
        \caption{Mvtec-AD (Tile)}
        \label{apd:caserea-5}
    \end{subfigure}
    \hfill
    \begin{subfigure}[b]{0.48\linewidth}
        \centering
        \includegraphics[width=\linewidth]{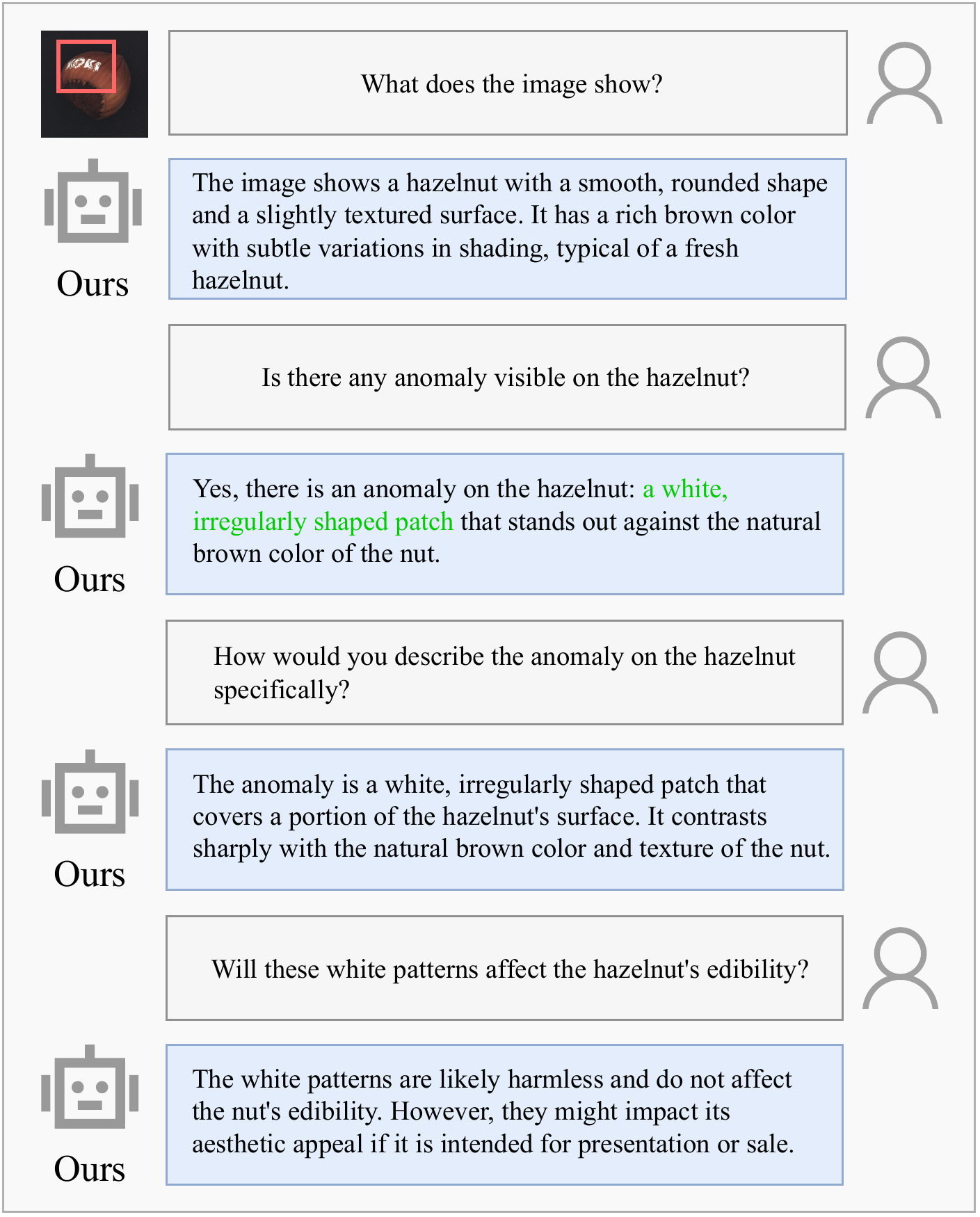}
        \caption{Mvtec-AD (Hazelnut)}
        \label{apd:caserea-6}
    \end{subfigure}

    \vspace{2mm}

    \begin{subfigure}[b]{0.48\linewidth}
        \centering
        \includegraphics[width=\linewidth]{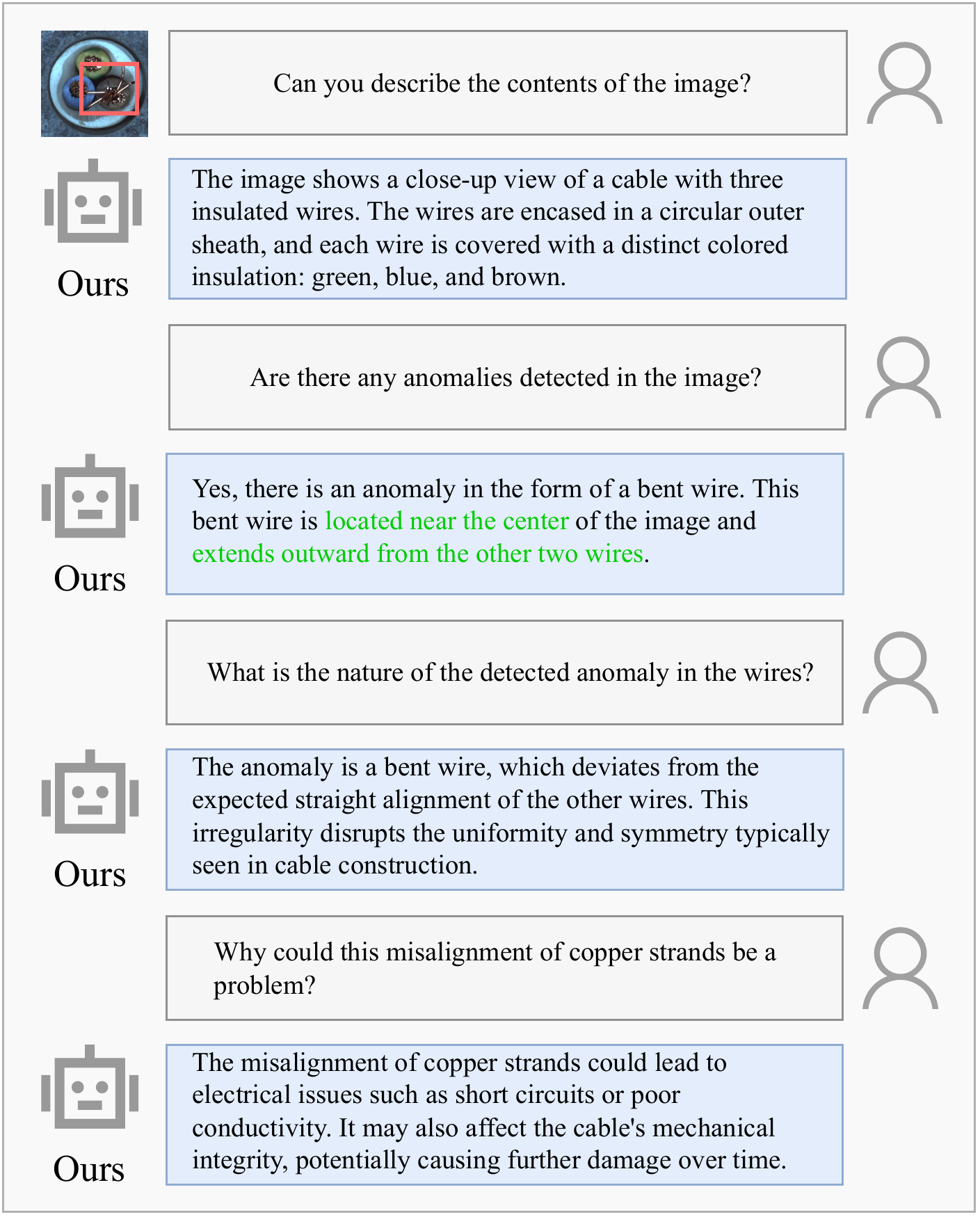}
        \caption{Mvtec-AD (Cable)}
        \label{apd:caserea-7}
    \end{subfigure}
    \hfill
    \begin{subfigure}[b]{0.48\linewidth}
        \centering
        \includegraphics[width=\linewidth]{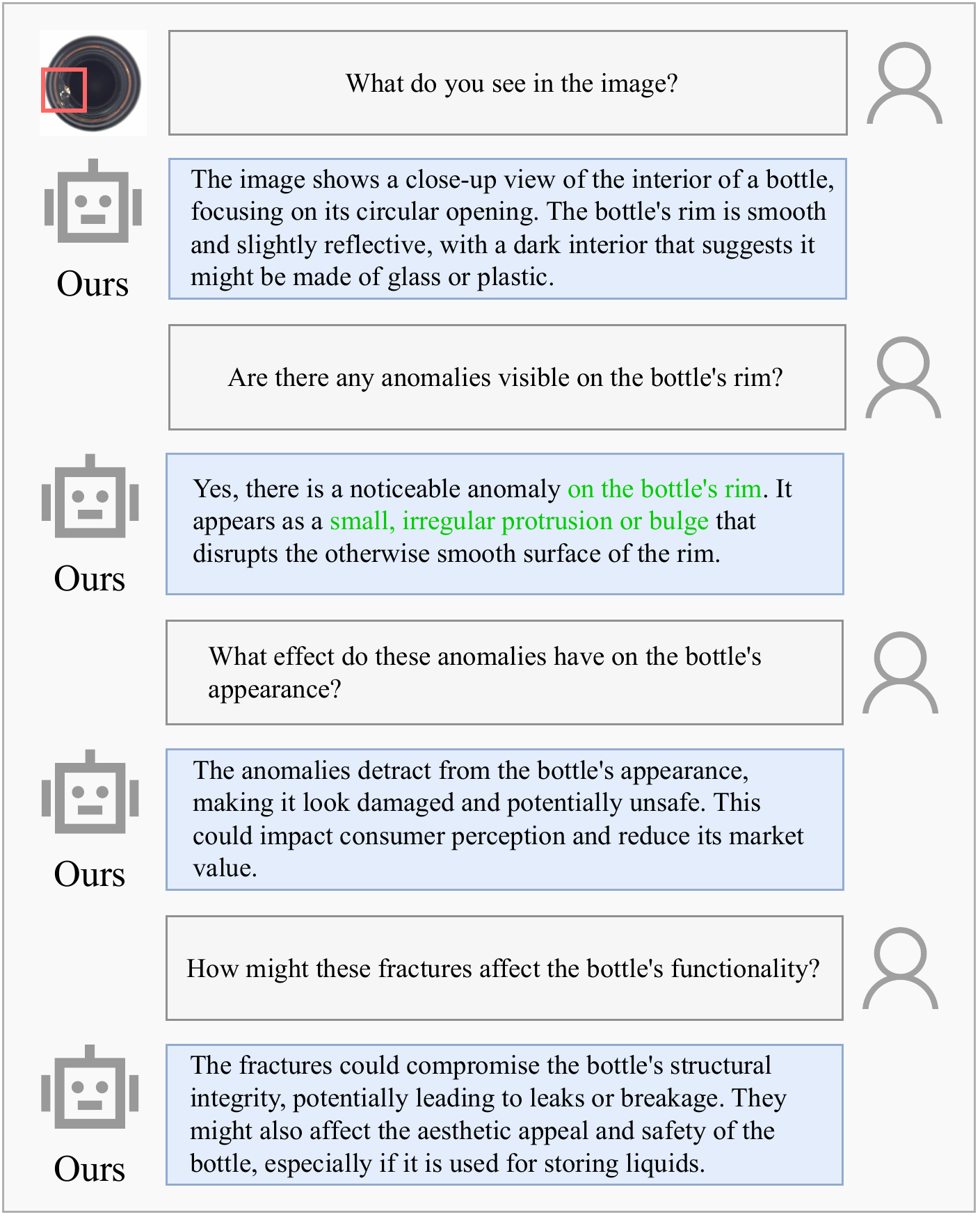}
        \caption{Mvtec-AD (Bottle)}
        \label{apd:caserea-8}
    \end{subfigure}

    \caption{Anomaly reasoning examples across datasets (part 2)}
    \label{fig:caserea_page2}
\end{figure*}

\noindent\textbf{Visualization Examples of Anomaly Reasoning. }
We further provide additional visualizations of anomaly detection and reasoning (\cref{apd:caserea-1}-\cref{apd:caserea-8}). CL-Anomaly goes beyond merely detecting anomalies in images, exhibiting robust reasoning capabilities that accurately address a broad range of anomaly-related queries. Moreover, as a single unified model capable of multi round human–machine interaction and handling anomaly question-answering across diverse domains and modalities,  CL-Anomaly provides a significant advantage over traditional anomaly detection approaches.

